\documentclass[lettersize,journal]{IEEEtran}
\usepackage{amsmath,amsfonts}
\usepackage{algorithmic}
\usepackage{algorithm}
\usepackage{array}
\usepackage[caption=false,font=normalsize,labelfont=sf,textfont=sf]{subfig}
\usepackage{textcomp}
\usepackage{stfloats}
\usepackage{url}
\usepackage{verbatim}
\usepackage{graphicx}
\usepackage[table]{xcolor}
\usepackage{cite}
\usepackage{multirow}
\usepackage{booktabs}
\usepackage[nottoc]{tocbibind}

\usepackage[hidelinks]{hyperref}

\begin{document}

\title{Weather-Magician: Reconstruction and Rendering Framework for 4D Weather Synthesis In Real Time}

\author{Chen Sang, Yeqiang Qian*, Jiale Zhang, Chunxiang Wang and Ming Yang*
\thanks{Chen Sang is with the SJTU Paris Elite Institute of Technology, Shanghai Jiao Tong University, Shanghai, 200240, China.

Yeqiang Qian, Jiale Zhang, Chunxiang Wang and Ming Yang are with the Department of Automation, Key Laboratory of System Control and Information Processing, Ministry of Education of China, Shanghai Jiao Tong University, Shanghai 200240, China (e-mail: qianyeqiang@sjtu.edu.cn; mingyang@sjtu.edu.cn) Corresponding authors*: Yeqiang Qian; Ming Yang.

This work is supported by the National Natural Science Foundation of China under Grant 62473253.}}



\maketitle

\begin{figure*}[!t]
    \centering
    \includegraphics[width=0.9\textwidth]{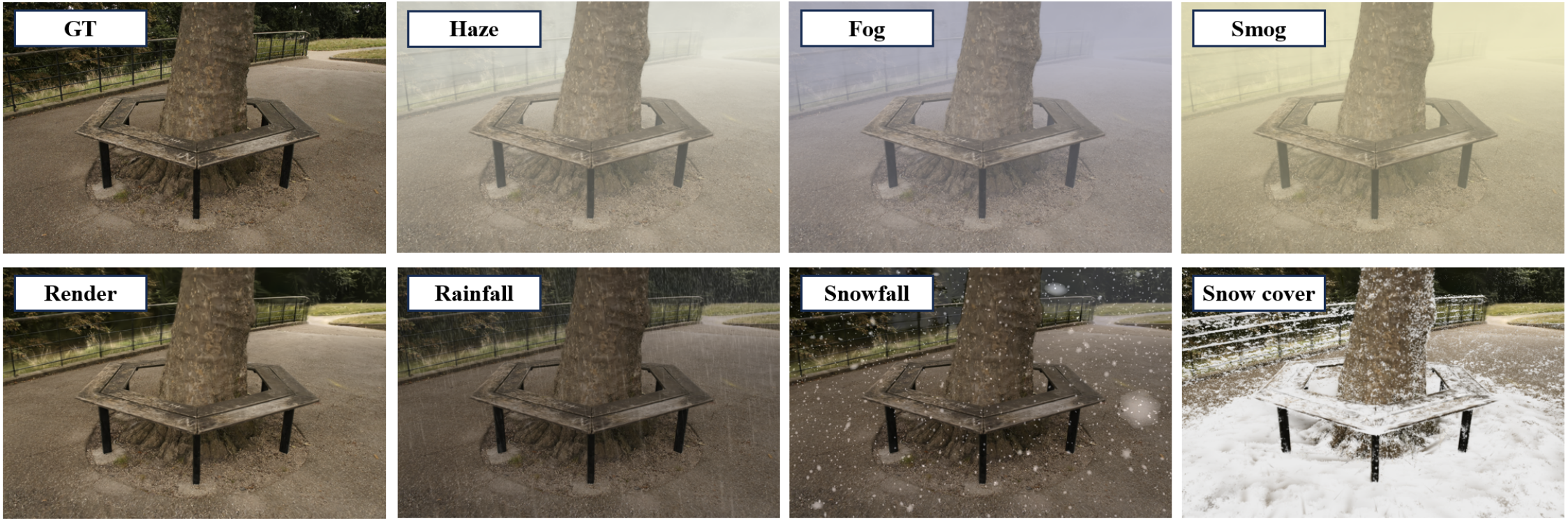}
    \caption{\textbf{Overall weather synthesis.} Weather-Magician, based on the vanilla 3DGS, utilizes the rendered depth, normal map, combined with the Gaussian modeling of raindrops, snowflakes, and accumulated snow, to achieve diverse weather simulation and rendering on reconstructed scenes. Our work not only exhibits high-fidelity synthesis results but also delivers real-time rendering performance.
}
    \label{overall}
\end{figure*}

\begin{abstract}
For tasks such as urban digital twins, VR/AR/game scene design, or creating synthetic films, the traditional industrial approach often involves manually modeling scenes and using various rendering engines to complete the rendering process. This approach typically requires high labor costs and hardware demands, and can result in poor quality when replicating complex real-world scenes. A more efficient approach is to use data from captured real-world scenes, then apply reconstruction and rendering algorithms to quickly recreate the authentic scene. However, current algorithms are unable to effectively reconstruct and render real-world weather effects. To address this, we propose a framework based on gaussian splatting, that can reconstruct real scenes and render them under synthesized 4D weather effects. Our work can simulate various common weather effects by applying Gaussians modeling and rendering techniques. It supports continuous dynamic weather changes and can easily control the details of the effects. Additionally, our work has low hardware requirements and achieves real-time rendering performance. The result demos can be accessed on our project homepage: \textcolor{blue}{\url{weathermagician.github.io}}

\end{abstract}


\section{Introduction}

\IEEEPARstart{D}{igital} environment modeling tasks have been emerging continuously in recent years. Most of them require the creation of 3D models from real scenes, which can be rendered into images for visualization. Traditional industrial approaches often involve using modeling platform such as Blender~\cite{blender} and 3ds Max, or simulators~\cite{dosovitskiy2017carla,shah2018airsim}, combined with rendering engines such as Eevee~\cite{blender}, Unity~\cite{noueihed2022knowledge}, and Unreal Engine~\cite{engine2018unreal} for scene modeling and rendering. These methods are widely adopted but frequently require manual modeling of entire scenes. In complex scenes, this process often faces several challenges: slow rendering speeds, high hardware demands, low quality manual modeling, and high labor and time costs. 

At the same time, Novel View Synthesis(NVS) has been of interest in the field of computer vision for decades, with its primary objective being to render a new view of a 3D scene from a sparse set of input views. Comparing to industrial methods, the automated reconstruction algorithms greatly boosts the efficiency of modeling tasks. The 3D reconstruction tasks include some traditional methods like \cite{schonberger2016structure,goesele2006multi}, and thanks to significant advancements in recent years, it becomes easier to reconstruct scenes and render high-quality images from novel viewpoints. The two predominant approaches that have emerged are Neural Radiance Fields(NeRF)~\cite{mildenhall2021nerf} and 3D Gaussian Splatting(3DGS)~\cite{kerbl20233d}. Compared to NeRF which relies on implicit representation and thus suffers from the performance impact of sampling, 3DGS demonstrates superior training and rendering speed with its explicit representation and rasterization method, which are undoubtedly major advantages that make it stand out.

However, a major challenge faced by current reconstruction algorithms, is their inability to effectively reproduce diverse real-world weather conditions, especially dynamic weather effects such as rain and snow. Reconstruction algorithms have ability to model dynamic objects in normal size\cite{wu20244d}, but are hard to handle with high-frequency noises like raindrops\cite{qian2024weathergs, liu2024deraings}. At the same time, the presence of weather conditions such as fog and haze can introduce numerous artifacts in the reconstruction outputs and significantly degrade the render quality. The explicit representation of 3DGS makes it relatively easy to perform editing operations~\cite{wang2024gaussianeditor,wang2025view,xu2025texture,rong2024gstex,chen2024dge}. However, most existing works fail to achieve complete and high-quality editing of weather effects\cite{dai2025rainygs,fiebelman2025let}. There are works\cite{wu2024gaussctrl, fiebelman2025let} integrating diffusion\cite{ho2020denoising} priors and NVS methods to perform the stylization of different weather types, but they only change the appearance of the original environment and haven't really create realistic weather elements. Also, diffusion-based methods are often hard to control the outputs, and may distort the real scene contents.

To address the problems, we first avoid directly reconstructing weathered scenes and instead, achieve weather synthesis by adding weather effects to clear-weather scenes. We designed a framework that integrates 3D reconstruction and the simulation of diverse weather conditions, enabling flexible control over scene weather effects. Our reconstruction method is based on the 3DGS~\cite{kerbl20233d} reconstruction and rendering framework, ensuring both realistic rendering quality and excellent performance. For the 4D weather simulation process, we create and edit the Gaussian, which is the fundamental element for gaussian splatting pipeline and is a simple explicit representation that facilitates manipulation and access to various elements of the modeled scene. We add new Gaussians to the original scene and edit their attributes according to weather types. Our framework supports the simulation and editing of static, dynamic, and cumulative weather effects. We leveraged the depth information of the reconstructed 3D scene to provide geometric support for static weather types like haze, fog, and smog. We have realized effective gaussian modeling of rainfall and snowfall, enabling the synthesis of these 4D weather elements while rendering them alongside the original scene. Those dynamic elements can show consistent movements among continuous frames, thus simulating falling effects. In addition, we have also provided methods to simulate snow accumulation within the scene. 

Fig. \textcolor{red}{\ref{overall}} shows the render results of different weather effects added to the same scene using our framework. The synthesis results are highly controllable. Our hardware-friendly framework achieves greater reality while maintaining real-time rendering performance. It also achieves view-consistency and time-consistency across different viewpoints, enabling the synthesis of dynamic weather-affected scene videos. In summary, our proposed method makes the following contributions:

\begin{itemize}
    \item We combine real-time, high-quality reconstruction and rendering techniques with high-quality 4D weather simulation. Our work shows high-fidelity synthesized results and performs well on consumer GPUs (e.g., Nvidia RTX 3060). 
    \item Our work supports the simulation of various common weather phenomena, such as fog, haze, smog, rain, and snow. The work also realizes realistic video synthesis with dynamic effects like rainfall and additional effects like accumulation of snow. Our weather simulation is highly controllable.
    \item Our work demonstrates realistic weather simulation performance across various scenes, whether images captured from small-scale environments or aerial footage from drones.
\end{itemize}

\section{Realted Works}
\subsection{\bf Industrial Modeling And Rendering Pipelines}
Traditional industrial pipelines often contain several processes: modeling, texturing, and rendering. For relatively simple scenes, modeling and texturing can be completed using techniques provided by simulators or rendering engines~\cite{dosovitskiy2017carla, shah2018airsim,noueihed2022knowledge,engine2018unreal}. Professional platforms like Blender~\cite{blender} and 3ds Max are more adaptable to complex scenes. Some rendering engines inside these platforms are applied to output results, such as Eevee, Cycles, Octane and Corona. Under certain pipelines, people can produce ideal models and create effects like weather. These processes requires extensive human intervention, resulting in high labor and time costs, and demands for the necessary artistic and technical skills. 

\subsection{\bf Traditional Weather Synthesis Methods}
To simulate weather effects, a simple method is to edit the original images to achieve a transformation from one original weather to another kind of weather. Traditional methods may use simple algorithms to generate the same additional weather layers and overlay them onto the original image, which lack generalization, often produce poor simulation effects, leading to bad simulation results. Some weather simulation methods, such as \cite{von2019simulating, nikolov2024digiweather}, model snow and fog upon different engines, then render and overlay the realistic weather effects onto the scene images. However, these methods rely solely on 2D image inputs and fail to effectively utilize the geometric information of the 3D scene, which can result in weather effects that lack realism.

\subsection{\bf Generative Methods}
The generative methods, such as GAN-based~\cite{goodfellow2020generative} and Diffusion-based~\cite{ho2020denoising} methods, enable the generation and repainting of high-quality images. ~\cite{chen2017stylebank, zhang2017style, wang2023stylediffusion} have effectively applied global style transformations to images and some of the extended works specifically focus on weather style transformations. For instance, Weather GAN~\cite{li2021weather} employs domain transfer to convert original images into target weather conditions. Another work integrate language models with generative models\cite{qian2024weatherdg}, enabling users to specify weather transformations via text prompts. However, most of these approaches tend to alter the content and structure of the original scene and cannot precisely control the details of the generated results. Since generative methods always operate on a single frame, applying them to different viewpoints of the same scene often results in inconsistent outputs, lacking view-consistency. Additionally, despite ongoing optimizations~\cite{ma2024deepcache, chen2023speed, lv2024fastercache}, they remain far from meeting real-time generation standards (e.g., 24 frames per second for films), making them unsuitable for applications requiring real-time visualization.

\subsection{\bf 3D Reconstruction And Weather Simulation}
~\cite{schonberger2016structure,goesele2006multi} are the traditional methods for performing sparse or dense 3d reconstruction tasks. The more recent Neural Radiance Fields (NeRF)~\cite{mildenhall2021nerf} and 3D Gaussian Splatting (3DGS)~\cite{kerbl20233d} can easily model real-world scenes into implicit or explicit representations significantly reducing the cost of modeling. 3DGS, in particular, leverages GPU-based rasterization rendering to achieve real-time scene rendering performance. However, existing 3D reconstruction algorithms often fail to reproduce weather effects present in the scene, such as rain, snowfall, and dense fog, which are all common weather conditions in various applications.

An approach to preserve weather effects is to simulate new weather on clear weather scenes. One method is to adding weather simulation modules before the render process, allowing for simultaneous rendering of both weather effects and the reconstructed scene. ClimateNeRF~\cite{li2023climatenerf} establishes a NeRF workflow that can render realistic weather effects such as smog, floods, and snow accumulation, but is unable to simulate and render dynamic weather effects and ensure real-time performance. \cite{chen2025styledstreets} uses diffusion priors to supervise the GS framework for stylizing reconstructed scenes. However, it can only change the appearance of the environment without introducing new and dynamic weather elements like raindrops. Rainy-GS\cite{dai2025rainygs} models realistic rainfall within 3DGS and simulates the behavior, splashes, and reflections of raindrops, while their raindrop particles themselves lack realism. \cite{fiebelman2025let} can simulate several static or dynamic weather effects and realize the interactions between the elements. However, its synthesized effects are rather unrealistic, and easily recognizable. In contrast, our proposed method can not only effectively simulate various common weather conditions but also achieves a good balance between real-time performance and realism.

\section{Methods}

\subsection{\bf Preliminaries}
\begin{figure*}[!t]
    \centering
    \includegraphics[width=0.85\textwidth]{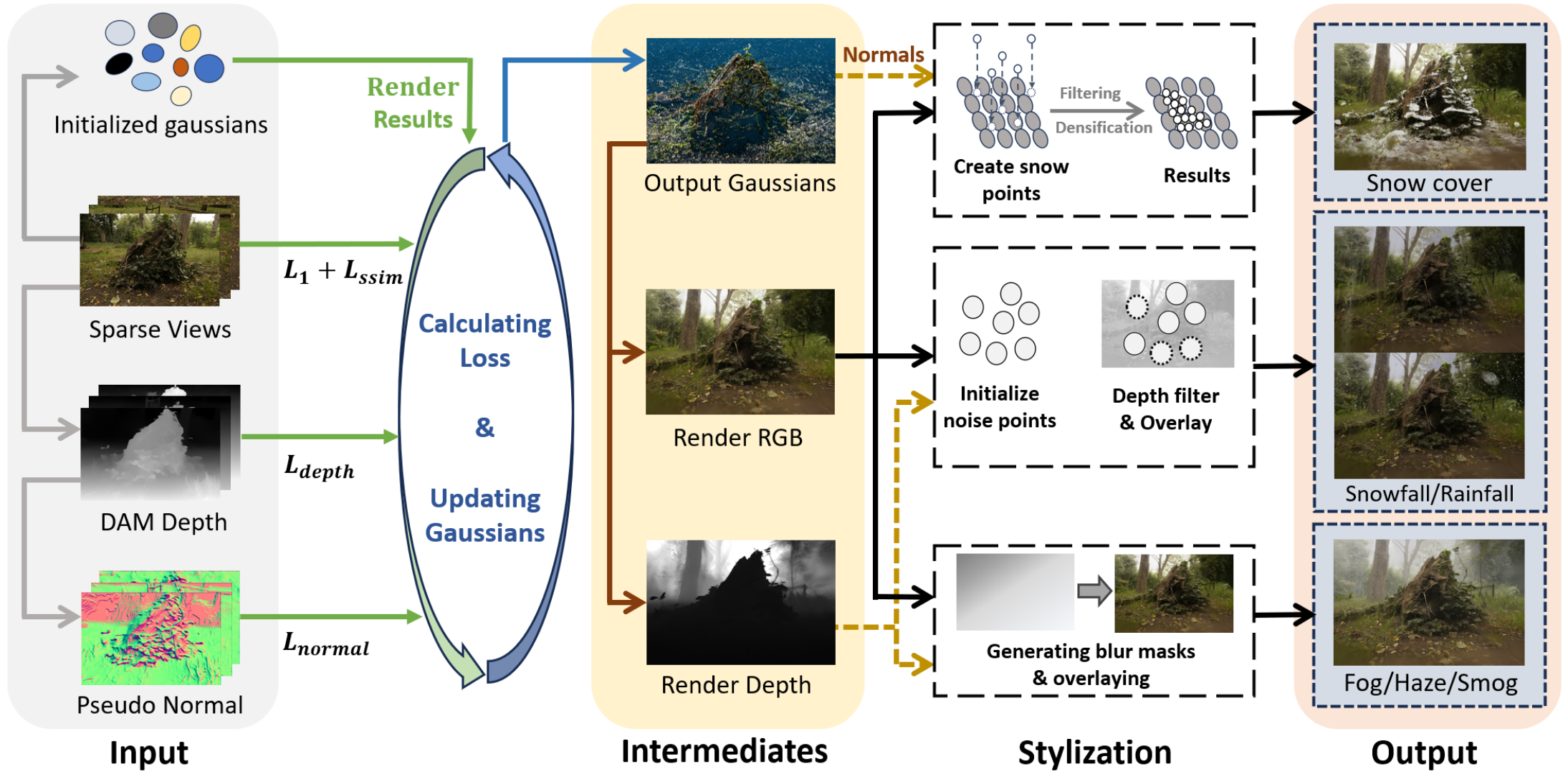}
    \caption{\textbf{Our weather synthesis framework.} Based on the traditional 3DGS training process, we incorporated depth and normal map supervision during training. Then, using the rendered RGB image, depth map, and the normal vector attributes of each Gaussian, the intermediates can enter different weather stylization processes to produce different simulation results.
}
    \label{fig framework}
\end{figure*}

3DGS\cite{kerbl20233d} explicitly represents an entire 3D scene using a large number of Gaussian distributions. Each Gaussian is defined as follows:  
\begin{align}
G(\mathbf{x}) = \exp\left(-\frac{1}{2} (\mathbf{x} - \boldsymbol{\mu})^\top \boldsymbol{\Sigma}^{-1} (\mathbf{x} - \boldsymbol{\mu}) \right),
\label{eq:def_gaussian}
\end{align}
where \(\boldsymbol{\mu}\) and \(\boldsymbol{\Sigma}\) represent the mean position and the 3D covariance matrix of each Gaussian, respectively. To ensure that the covariance matrix \(\boldsymbol{\Sigma}\) remains positive definite during the optimization process, it can be expressed as follows using a scale matrix \(\mathbf{S}\) and a rotation matrix \(\mathbf{R}\):  
\begin{align}
\boldsymbol{\Sigma} = \mathbf{R S S^\top R^\top},
\label{eq:def_sigma}
\end{align}
where \(\mathbf{S}\) and \(\mathbf{R}\) are derived from two additional attributes of each Gaussian: a scale vector \(\mathbf{s}\) and a quaternion \(\mathbf{q}\), which are stored and optimized during the training process.  

In addition, each Gaussian possesses two more attributes: opacity \(o\) which determines the weight of Gaussian point blending during the rendering process based on alpha blending, and color \(\mathbf{c}\), which is defined using spherical harmonics. The actual training process involves optimizing the five attributes of each Gaussian: \(\{\boldsymbol{\mu}, \mathbf{s}, \mathbf{q}, o, \mathbf{c}\}\).

The rendering process will first project Gaussians from 3D space onto the 2D image plane. The projected 2D Gaussian \(G'(x)\) on the image plane has a corresponding 2D covariance matrix $\boldsymbol{\Sigma}$:  
\begin{align}
\boldsymbol{\Sigma}' = \mathbf{JW} \boldsymbol{\Sigma} \mathbf{W^\top J^\top},
\label{eq:def_2dsigma}
\end{align}

where \(J\) is the Jacobian matrix of the linear approximation of the projection transformation, and \(W\) represents the coordinate transformation of the viewpoint.  

Thus, in the final rendered image, assuming the number of Gaussians projected onto a pixel \(p\) is \(N_p\), the pixel color \(C_p\) can be rendered using the following formula:  

\begin{align}
C_p = \sum_{i\in N_p} \alpha_i \mathbf{c}_i\sum_{j=1}^i (1-\alpha_j),
\label{eq:3dgs_render}
\end{align}

where \(\alpha_j = o_j G'(x)\) represents the transparency of the color blending. The blending of multiple Gaussians finally generates different colors.

\subsection{\bf Geometry Enhancements}

Our work is based on the vanilla 3DGS framework\cite{kerbl20233d} with slight modifications. As shown in Fig. \textcolor{red}{\ref{fig framework}}, our method starts with point clouds initialized with Structure from Motion(SfM)\cite{schonberger2016structure}, obtains intermediate results through the 3DGS training process, and renders different weather effects via various weather simulation modules. The reconstruction of real complex scene might introduce artifacts and result in a degraded geometric structure, leading to poor synthesis of weather effects. Therefore, we first manually adds an upper hemisphere point cover to the initialized point cloud, which can help assist in both sky RGB and depth rendering.Then we also use the depth and normal rendered by 3DGS to further optimize the Gaussian scene.


For the depth map, by extending the rasterization rendering framework of the original 3DGS, the absolute depth map \(D_{\text{abs}}\) can be directly obtained by weighted summation of the \(z\)-axis values of each Gaussian. To supervise depth, our work utilizes the Depth Anything Model (DAM)\cite{yang2024depth} to generate relative depth as pseudo ground truth for depth\(D_{\text{DAM}}\) for each training image. For the sky regions which may appear in outdoor scenes, we have assigned an additional depth offset to the sky pixels of \(D_{\text{DAM}}\) to further optimize the geometry and eliminate artifacts. During actual training, we set a maximum reference depth\(d_{\text{max}}\) to create a normalized reference depth map: \(D_{\text{ref}} = \min(1, D_{\text{abs}} / d_{\text{max}})\), where$D_{\text{abs}}$ is the output depth map rendered by 3DGS.

We have adopted the depth supervision loss from DNGaussian\cite{li2024dngaussian}, which uses Hard and Soft Depth Regularization to separately optimize the positions and opacities of Gaussians step by step. Additionally, Global-Local Depth Normalization is incorporated to enhance depth details. The $\mathcal{L}_{hard}$ and $\mathcal{L}_{soft}$ terms can be unified and expressed as follows\cite{li2024dngaussian}:
\begin{align}
\mathcal{L}_{\text{hard/soft}} = || D_{ref}^{GN}-D_{DAM}^{GN} ||_2 + \gamma|| D_{ref}^{LN}-D_{DAM}^{LN} ||_2,
\label{eq:hs depth loss}
\end{align}

Here, $GN$ represents the use of global normalization, and $LN$ represents local normalization. For more specific details, please refer to the original paper \cite{li2024dngaussian}. Thus, the final depth loss $\mathcal{L}_{\text{depth}}$ is defined as:
\begin{align}
\mathcal{L}_{\text{depth}} = \mathcal{L}_{hard} + \mathcal{L}_{soft},
\label{eq:depth loss}
\end{align}

To supervise the training of normals, we first extract a pseudo-normal ground truth \(N_{\text{pseudo}}\) from \(D_{\text{DAM}}\). Directly adding a normal attribute to Gaussians can result in difficulties in convergence during training, and the added attributes are decoupled from the covariance, offering little assistance in maintaining geometric structure. Therefore, we directly assign the shortest axis of each Gaussian as its corresponding normal vector. Finally, the normal consistency loss is the $L_2$ regularization loss between the pseudo-normal map \(N_{\text{pseudo}}\) and the rendered normal map \(N\):

\begin{align}
\mathcal L_{normal}=|| N-N_{pseudo} ||_2,
\label{eq:normal l2_loss}
\end{align}

The final loss for training process can be expressed as:
\begin{align}
\mathcal{L}_{\text{total}} = \mathcal{L}_{1} + \mathcal{L}_{depth} + \lambda_{ssim} \mathcal{L}_{ssim} + \lambda_{normal} \mathcal{L}_{normal},
\label{eq:total loss}
\end{align}
where $\mathcal{L}_{1}$ and $\mathcal{L}_{ssim}$ represent the $L_1$ loss and structural similarity (SSIM) loss between the rendered image and ground truth image, as used in the original 3DGS.

To prevent normal consistency loss from negatively optimizing the opacity and causing rendered images to exhibit object transparency, the optimization of $\mathcal{L}_{normal}$ is delayed for $it_{N}$ iterations during training. This allows $\mathcal{L}_{depth}$ to sufficiently optimize the positions and opacities of the Gaussians before introducing $\mathcal{L}_{normal}$ into the training process. Then the normal map is rendered using the standard opacity blending strategy. In the experiments presented in this work, $it_{N}$ is set to 6000 iterations.

\subsection{\bf Static Weather Effects Simulation}
 In weather phenomena such as fog, haze, and smog, their shared characteristic is the blurring effect of distant views, while they differ mainly from each other in the size, density, and scattering color of the floating particles. We define these blurring effects as static weather because they usually do not show great visual changes during small time periods(one or two minutes). To simplify and unify the explanation of such blurring effects, we will focus on fog as the representative case. Since our method supports highly customizable parameters, the other phenomena can be simulated by adjusting the color or blurring intensity. 
For synthesizing static weather, the primary goal is to simulate blurry and hazy effects. \cite{li2023climatenerf} achieves the smog effect by modifying the volumetric density to fill the entire space. Simulating fog by adding Gaussians leads to difficulties to determine the attributes, blending color errors and increase computational burden. Instead, we extract a blurring mask from the rendered depth map \(D_\text{ref}\) and overlap it to the rendered RGB \(C_\text{render}\). Assuming the final sky color in the distance under foggy conditions is \(C_\text{fog}\), the final RGB color of the rendered image can be expressed as follows:  
\begin{align}
C_{render,fog}=C_{fog}\times \alpha_{style}+C_{render}\times(1-\alpha_{style})
\label{eq:fog style}
\end{align}
Here, \(\alpha_\text{style}\) represents the depth-based fog coefficient. We assume that particles are uniformly distributed in space, leading to visibility following an exponential relationship inversely proportional to depth. If we predefine a parameter \(I_\text{style}\) to control the global intensity of the effect, \(\alpha_\text{style}\) can be computed as:  
\begin{align}
\alpha_\text{style} = \min(1, 1-\exp(-I_\text{style} \cdot D_\text{ref})),
\label{eq:style_intensity}
\end{align}

With this straightforward post-processing approach, it is possible to apply the blurring effect to the rendered image in real time. Moreover, since \(C_\text{fog}\) and \(I_\text{style}\) can be manually adjusted, blurring effects with different colors, such as haze and smog, can also be simulated in the same way.

\subsection{\bf Dynamic Weather Effects Simulation}

\begin{figure*}[!t]
    \centering
    \subfloat[Original]{
        \includegraphics[height=1.65in]{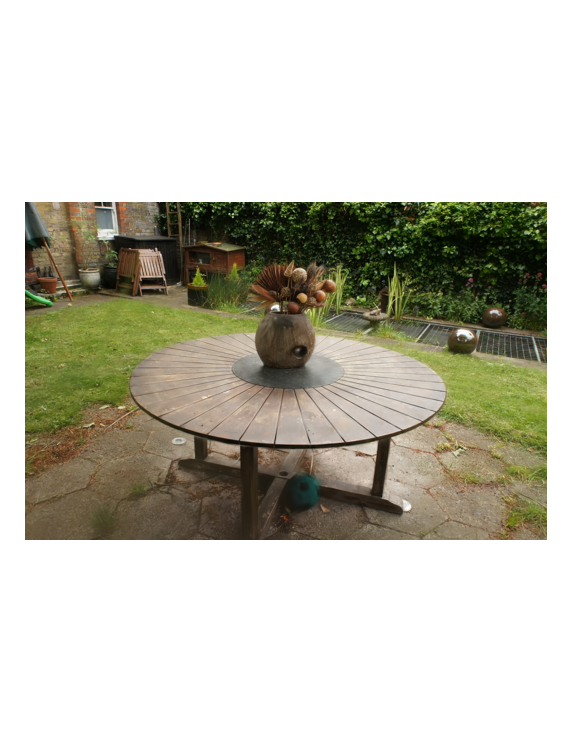}
        \label{fig_clean}
    }
    \hfill 
    \subfloat[Noise Scene]{
        \includegraphics[height=1.65in]{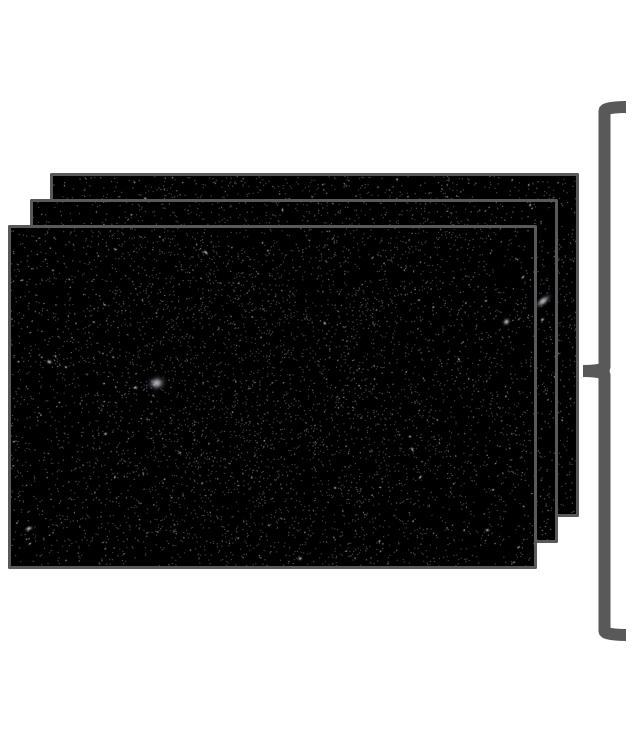}
        \label{fig_noise}
    }
    \hfill
    \subfloat[Overlay Result]{
        \includegraphics[height=1.65in]{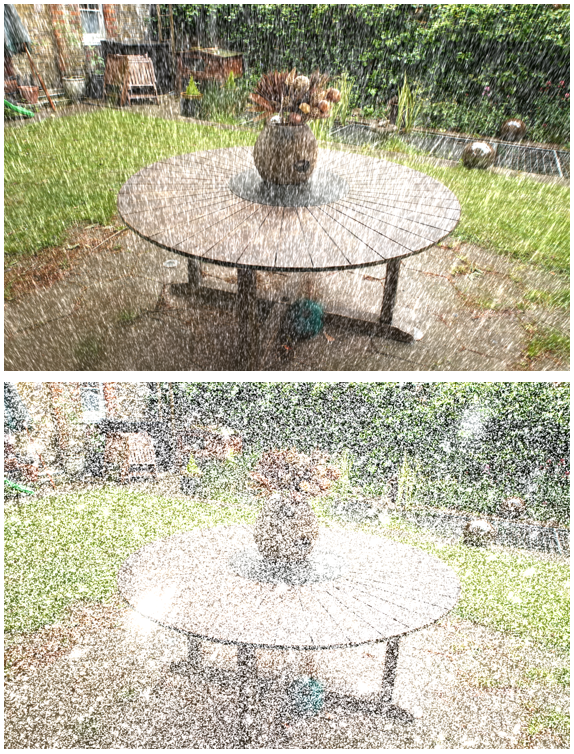}
        \label{fig_direct}
    }
    \hfill
    \subfloat[Filtered Scene]{
        \includegraphics[height=1.65in]{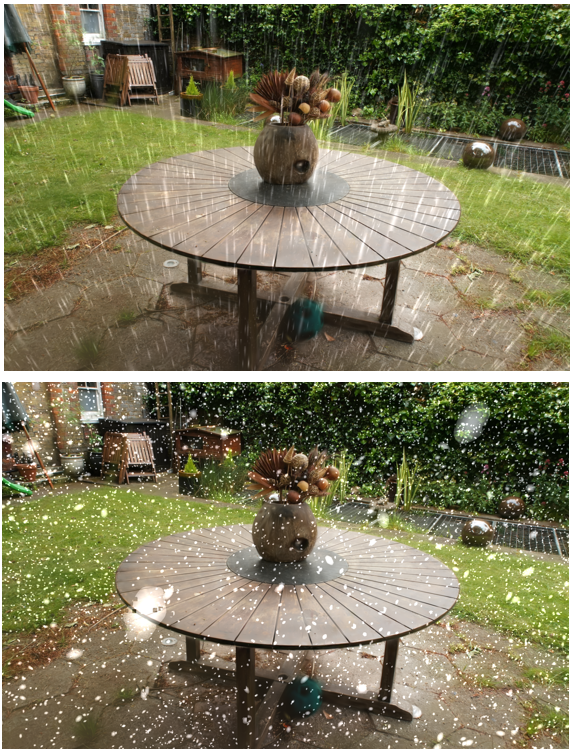}
        \label{fig_filtered}
    }
    \hfill
    \subfloat[Final Effect]{
        \includegraphics[height=1.65in]{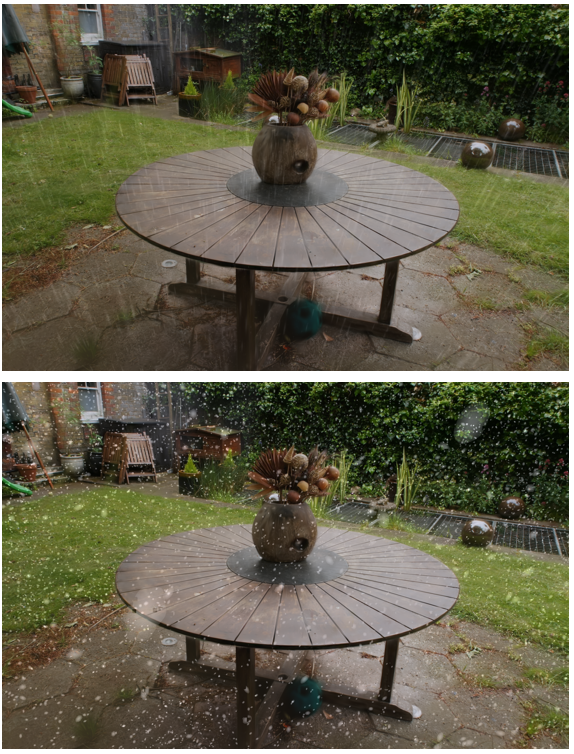}
        \label{fig_final}
    }
    \caption{\textbf{Our pipeline with different stages to simulate raindrops and snowflakes effect.} The first line simulates the raindrops, and the second line simulates the snowflakes. First we generate several noise layers to represent noise Gaussian scene. (c) shows the direct overlay results of the two scenes, which demonstrates great unreality. So we filter the occluded Gaussians by comparing their depth values. Finally, these noises are overlaid onto the original image with transparency determined by comparing scene brightness. Additionally, we applied simple stylization to the output image, such as color adjustments and blurring.
}
    \label{fig_pipeline}
\end{figure*}
Rainfall and snowfall can generally be classified into the category of dynamic weather, where dynamic entities may appear, like raindrops or snowflakes. The main objective of both is to simulate and visualize 4D raindrops or snowflakes, which are essentially floating noise particles. For the modeling of each floating noise particle, an explicit Gaussian representation is highly beneficial. A simple approach is to bind a Gaussian to each noise particle and randomly populate the scene with these Gaussians for initialization. The overall pipeline of noise particles simulation is shown in Fig. \textcolor{red}{\ref{fig_pipeline}}. The following parts will explain this process in detail.

\begin{figure}[ht]
    \centering
    \subfloat[Snowflake]{
        \includegraphics[width=0.15\textwidth]{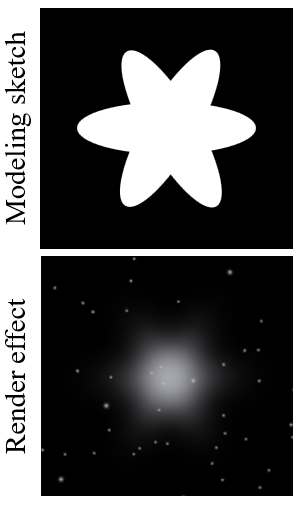}
        \label{fig_model_snow}
    }
    \subfloat[Raindrop]{
        \includegraphics[width=0.15\textwidth]{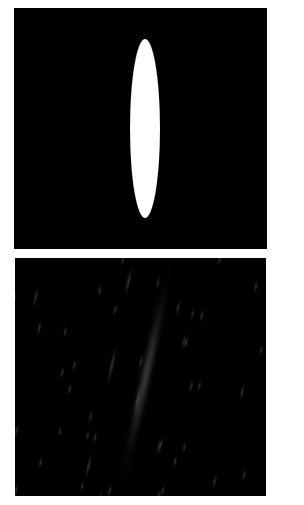}
        \label{fig_model_rain}
    }
    \caption{\textbf{Modeling sketch and actual render results of raindrops and snowflakes.} The first row shows how we design each snowflake and raindrop with gaussians. The second row represents the actual rendering result.
}
    \label{fig_model}
\end{figure}

Fig. \textcolor{red}{\ref{fig_model}} shows how our method models each raindrop and snowflake. For raindrops, they often exhibit vertical stretching in the direction of fall. Therefore, an elongated Gaussian is enough to model a raindrop. At the same time, snowflakes tend to maintain their original crystalline shape. In this case, our work models a snowflake using three elongated Gaussians with angles of $\frac{\pi}{3}$ among them. During initialization, snowflakes and raindrops are assigned random rotations. Their initial colors are set to a grayish RGB tone. 


After initializing the noise particles, the insertion of those new Gaussians to the scene may lead to issues such as background color changes or wrong occlusion due to the blending process of 3DGS. Therefore, this work adopts a method for separating the original Gaussian scene $G$ and the Gaussian scene of noise particles $G_n$. $G$ and $G_n$ are rendered separately to produce the scene image and pure noise mask, respectively, and then the two images are weighted and combined to output the final RGB image of rain or snow $C_{fall}$. In addition, our work also introduces a brightness-based transparency factor, which ensures that the noise particle colors do not stand out excessively, thus enhancing the realism of the simulation. The final formula for combining the noise and scene images is as follows:
\begin{align}
C_{fall} = f_l C_{noise}+C_{render},
\label{eq:combining noise masks}
\end{align}
where $f_l$ is the luminance blending factor, calculated by comparing the brightness $L^p$ of pixel $p$ in the rendered image with the average sky brightness $L_{sky}$, $L^p$ is obtained by the sum of average weighted 3-channel colors:
\begin{align}
L^{p} = \frac{C_{render,R}^p+C_{render,G}^p+C_{render,B}^p}{3},
\label{eq:computing pixel luminance}
\end{align}
\begin{align}
L_{sky}=\frac{1}{N_p}\sum_{p\in sky}L^p,
\label{eq:computing sky luminance}
\end{align}
So the $f_l$ at each $p$ will be:
\begin{align}
f_{l}^p=\exp(\min(\max(L_{sky}-L^p,t_{max}),t_{min}))-1,
\label{eq:computing luminance blending factor}
\end{align}
where $t_{max}$ and $t_{min}$ are manually set parameters. $t_{min}$ can be set to a negative value(e.g., -0.1 for snowflakes) to generate a dim backlight effect. A moderate $t_{max}$ can limit the maximum luminance of the noise, thus creating realistic effects. By collecting all $f_{l}^p$, the final overlay operation of the two image layers can be achieved.


For further enhancing the synthesis quality, a distance threshold $t_D$ is set to filter out the distant noise points, as they tend to occupy only a single pixel after rendering, reducing realism. To deal with the case of occlusion, we generate a mask based on the depth comparison to filter the occluded noise points. The depth map of the noise layer $D_{ref,n}$ is thus rendered and scaled during the process We also set a color intensity threshold $t_C$ to sharpen the edges of the noises. Thus, the mask $m$ is calculated as:
\begin{align}
m=(D_{ref,n}<D_{ref})\cap(D_{ref,n}<t_D )\cap(C_{noise}>t_C ),
\label{eq:noise overlay mask}
\end{align}


However, there is an issue with directly comparing depth maps to filter occluded noise points: Suppose there are two noise Gaussians at different distances from the camera, both projected onto the same pixel. The mask filtering the further noise may erroneously filter the near noise as well. To address this, we decompose the noise Gaussian point cloud $G_n$ into $k$ sub-Gaussian point clouds $G_n^i,i\in[0,k]$, each of which contains a certain maximum number of Gaussians. This approach significantly reduces the occurrence of multiple overlapping noise points at the same pixel in each sub-noise layer rendered. Therefore, the final rain and snow simulation process can be summarized as the weighted superposition of the background layer $C_{render}$ rendered by 
$G$, combined with the mask processing of the noise sub-layers $C_{noise}^i$ rendered by $G_n^i$:
\begin{align}
C_{fall}=\sum_{i}^{k}f_l^i\times (C_{noise}^i \in m^i)+C_{render},
\label{eq:sub-layers overlay}
\end{align}
where $f_l^i$ is the different $f_l$ calculated with different noise sub-layers.

In continuous frames, we can add different displacement to each noise gaussian to simulate the falling process in videos. Additionally, in scenes with heavy rainfall or snowfall, it is often observed that the distant scenery appears blurry and hazy. We can also add some light fog effect generated in part III.C, to simulate the phenomenon in real world.


\subsection{\bf Cumulative Weather Effect Simulation}
\begin{figure}[ht]
\centering
\subfloat[Original]{
    \includegraphics[height=0.7in]{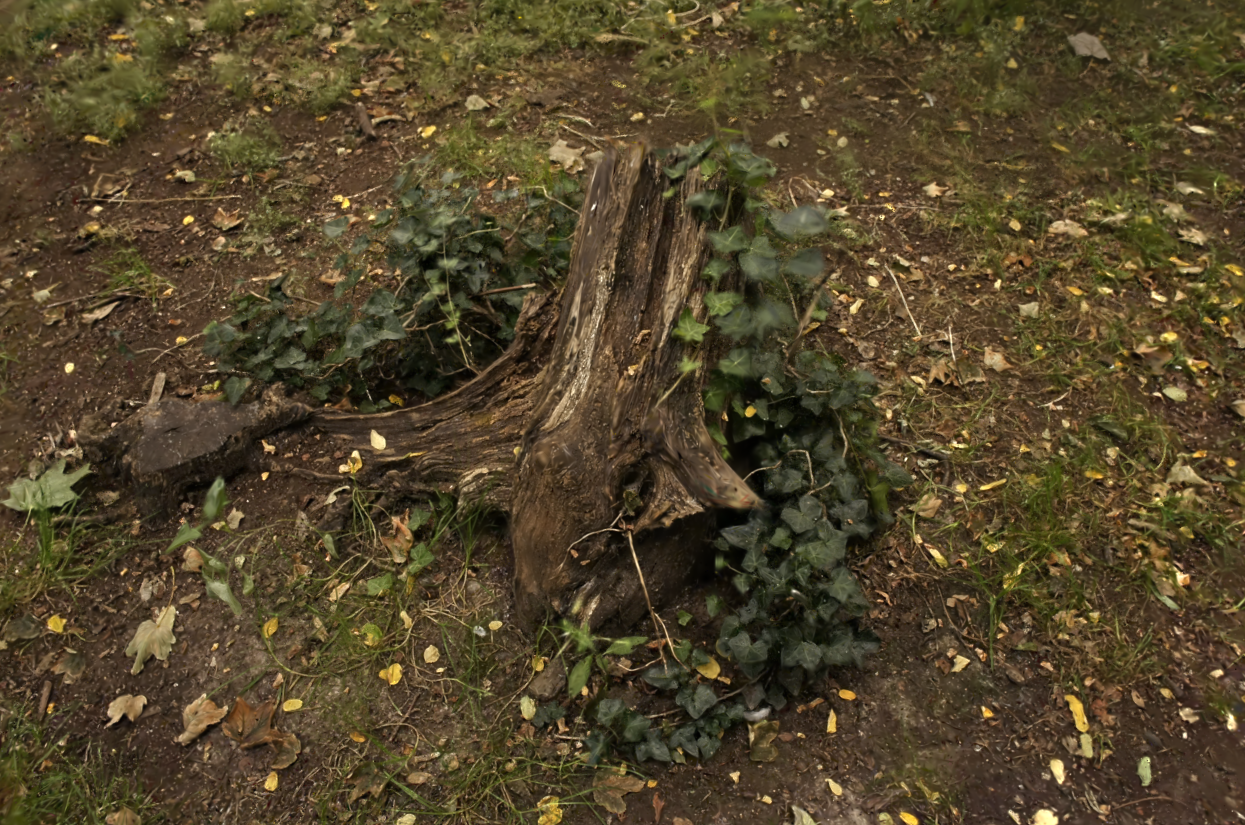}
    \label{fig_sc_clean}
}
\subfloat[Initialization]{
    \includegraphics[height=0.7in]{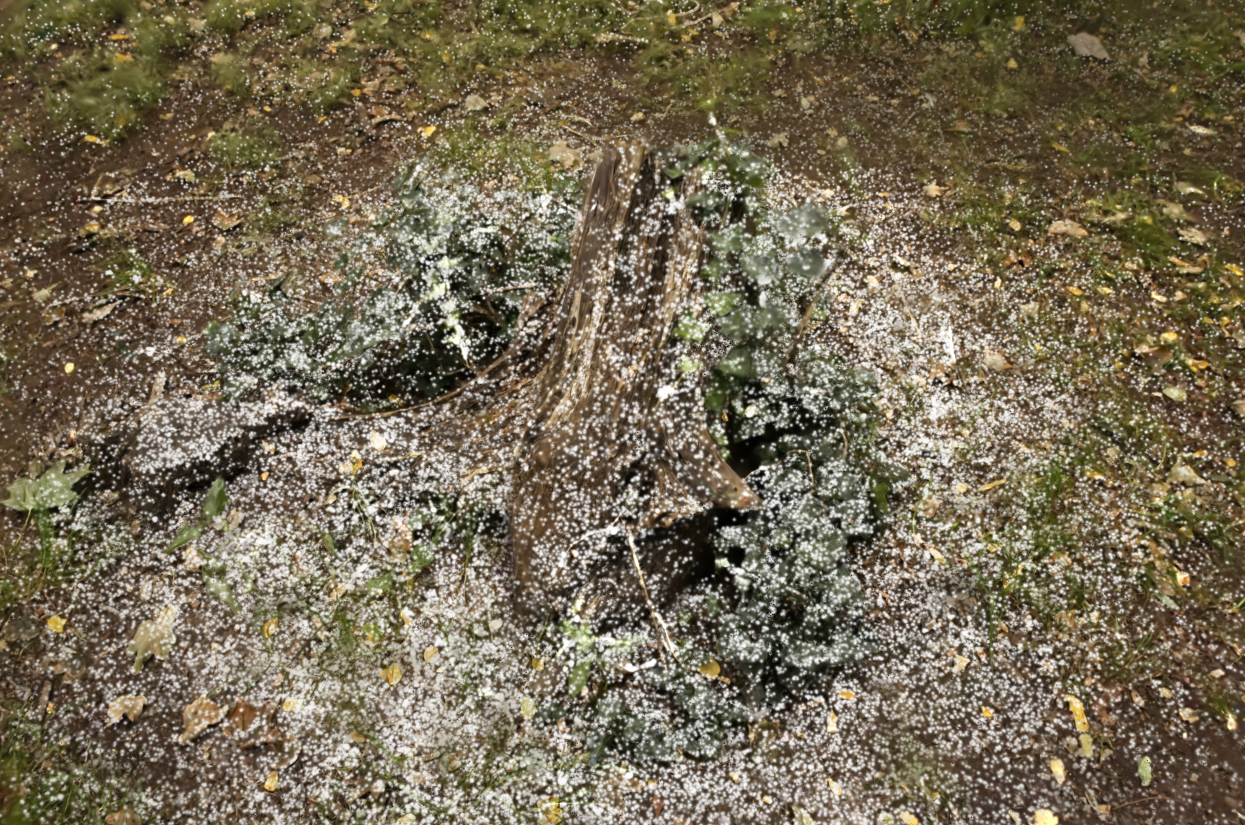}
    \label{fig_sc_init}
}
\subfloat[Final result]{
    \includegraphics[height=0.7in]{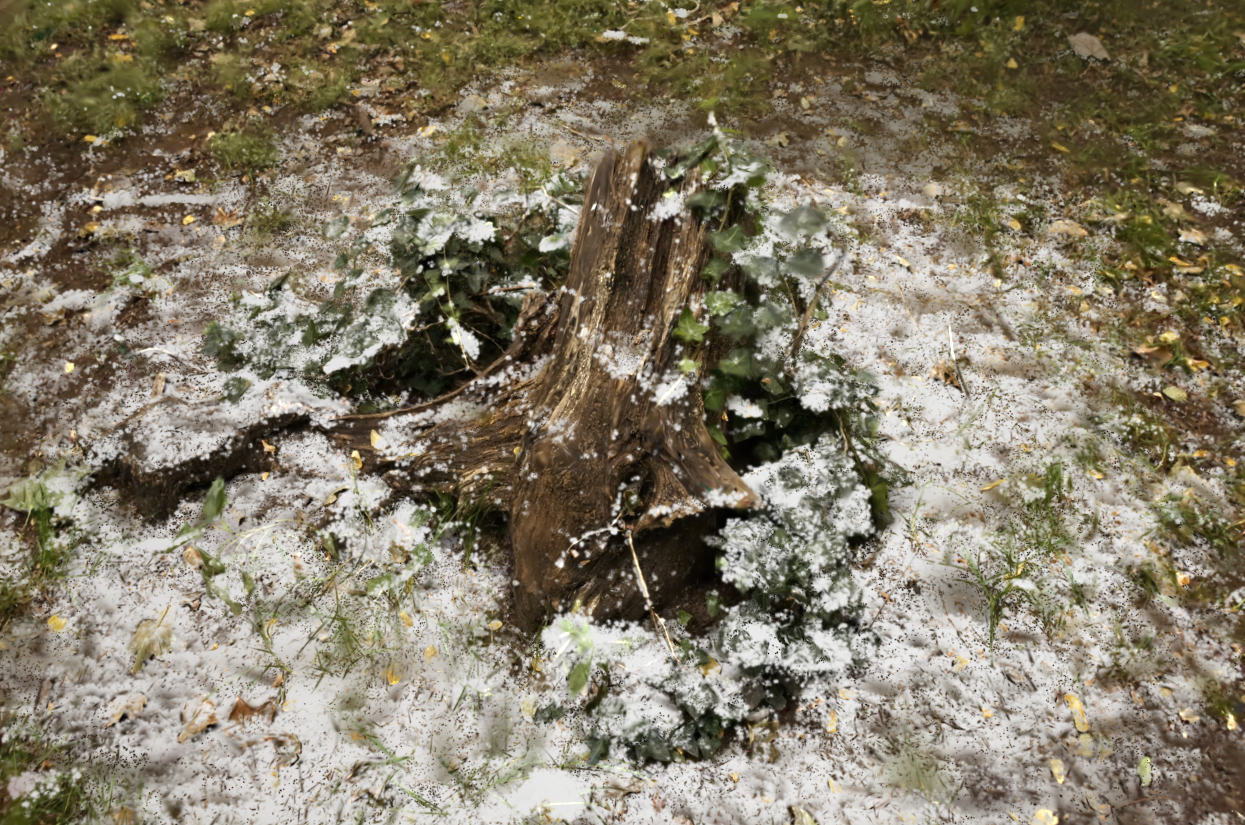}
    \label{fig_sc_final}
}
\caption{\textbf{Snow accumulation simulation.} The snow position are first initialized in the scene, like what \ref{fig_sc_init} shows. Then the local plane densification will be performed to simulate snow clusters. After densifying the snow, the remained outliers will be filtered. The final effect is shown in \ref{fig_sc_final}.
}
\label{fig_sc}
\end{figure}

To support the simulation of the varied environment appearance under certain cumulative weather effects, we here provide the method to simulate the weather phenomenon of snow accumulation. Fig. \textcolor{red}{\ref{fig_sc}} shows the qualitative result of each snow cover simulation process. First, for each Gaussian $g$, we calculate its normal vector $\mathbf{n}_g$ by the shortest axis direction. By taking the dot product of $\mathbf{n}_g$ and the negative gravity vector $\mathbf{g}$, and setting a minimum initialization threshold, Gaussians with upward-facing normals can be chosen as the specific locations of the snow.

Gaussians may be sparse depending on the scene, so interpolation is needed to ensure the realism of snow in the final rendering. In nature, snow often appears as smooth planes. Therefore, local plane estimation can be used after initialization to extract the plane on which each snow Gaussian resides. The supervision of normal maps during training ensures the feasibility of this method. For a initialized snow Gaussian $p_{\text{snow}}$, given $k$ nearest neighbors, the proposed plane radius $\mathbf{r}_{plane}$ is defined as:
\begin{align}
\mathbf{r}_{plane}=\frac{\text{median}(R_n)}{1 + 2\boldsymbol{\sigma}_n},
\label{eq:plane radius}
\end{align}
where $R_n$ is the set of distances from the nearest neighbors to $p_{\text{snow}}$, and $\boldsymbol{\sigma}_n$ is the standard deviation of $R_n$. Random points are filled on the plane using a uniform probability within $\mathbf{r}_{plane}$. This plane radius adapts well to local point densities automatically. 

Furthermore, the prerequisite for generating planes is that the angle between the plane’s normal vector and the gravity vector is less than $\frac{\pi}{6}$, to avoid bad interpolation. We filter the outlier snow Gaussian at the end of the process. Finally, by adjusting the Gaussian scale, opacity, and color appropriately, the scene with snow cover added can be rendered in real-time while ensuring view-consistency. Regarding the previously mentioned issues of gaussian blending, since the snow is generally expected to fully cover the region it occupies, showing the color of the snow itself, increasing its $\alpha$ can highlight its color and prevent transparent errors. However, this sometimes leads to wrong display of snow behind objects, which can be a direction for future optimization.

\section{Experiments}

\subsection{\bf Datasets}
Our work is tested on multiple datasets: we trained and performed weather editing works on several outdoor scenes from Mip-NeRF360\cite{barron2022mip}, such as \textit{Stump}, \textit{Garden}, \textit{Treehill} and \textit{Bicycle}. Additionally, we manually collected several traffic scenes at Shanghai Jiao Tong University (SJTU), consisting of thousands of images captured from a single forward-facing camera mounted on a car. Also, we reconstructed a scene of SJTU's main gate captured via aerial drone footage, which consists of about six thousand images extracted from a aerial video. These open outdoor scenes provide essential conditions for weather simulation, with varying scene scales and rich testing significance.

For performing the quantitative experiments, we primarily used a series of real weather datasets including the Dense\cite{Bijelic_2020_CVPR} dataset (12,997 frames, containing rain, snow, fog and smog weather types), the SMOKE dataset\cite{jin2022structure} (116 frames, containing only artificial smoke), the RTTS dataset\cite{li2019benchmarking} (4,322 frames of haze weather), and the RRID dataset\cite{li2019single} (34,915 frames of rainy weather, from various datasets including real rain scenes and images from the Internet) for testing. Based on manual qualitative evaluation and the split files provided by the datasets, we divided images with different weather types, ultimately forming the real weather dataset for evaluation (5,643 frames of foggy weather, 3,258 frames of rainy weather, and 4,623 frames of snowy weather).

\subsection{\bf Baselines}
Relevant works in the generative domain include state-of-the-art(SOTA) methods like \textbf{ClimateGAN\cite{schmidt2021climategan}} and \textbf{Stable Diffusion (SD)\cite{rombach2022high}}, where ClimateGAN is capable of high-quality smog simulation, and SD can simulate various weather conditions based on different text prompts. Additionally, \textbf{ClimateNeRF\cite{li2023climatenerf}} serves as a SOTA method for 3D reconstruction and rendering combined with weather editing. It supports smog and snow accumulation simulation and provides view-consistency render results.

\subsection{\bf Experiment settings}
All training and render results for this work were conducted on a single NVIDIA GeForce RTX 4090 GPU. Since the quality of weather simulation results heavily relies on subjective evaluation, we will mainly present qualitative experiment results to verify the high-quality simulation results. In addition, we provide some quantitative results as a reference for performance evaluation.

\subsection{\bf Results on all the datasets}
\begin{figure*}[t!]
    \centering

    \subfloat[Original]{
        \includegraphics[width=0.14\textwidth]{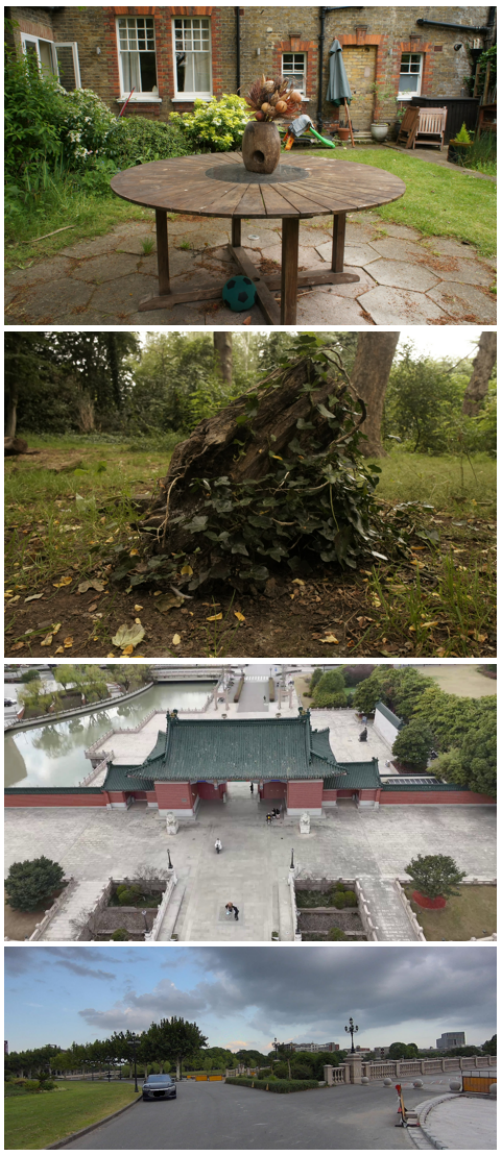}
        \label{fig_dataset_clean}
    }
    \hspace{-3mm} 
    \subfloat[Haze]{
        \includegraphics[width=0.14\textwidth]{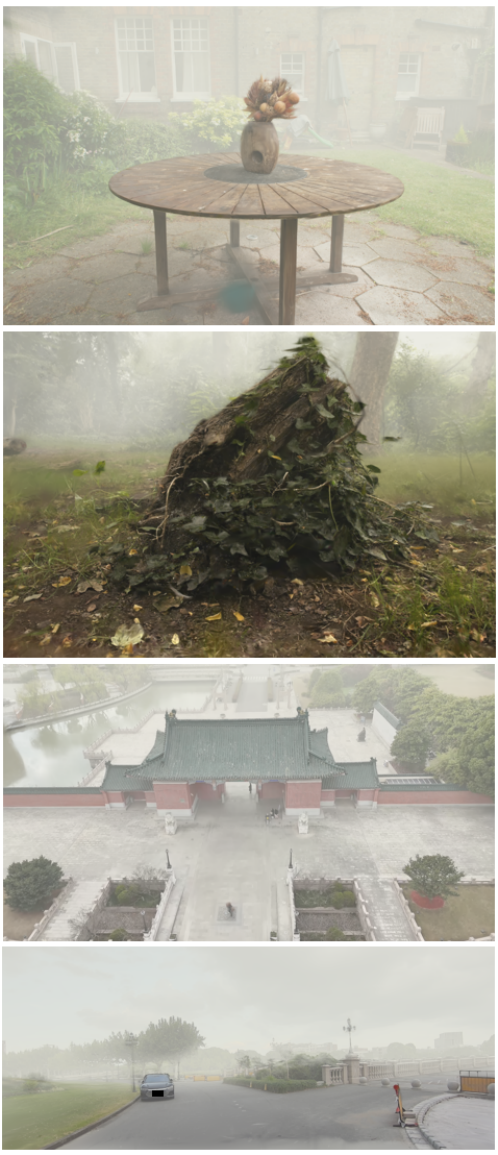}
        \label{fig_dataset_haze}
    }
    \hspace{-3mm}
    \subfloat[Fog]{
        \includegraphics[width=0.14\textwidth]{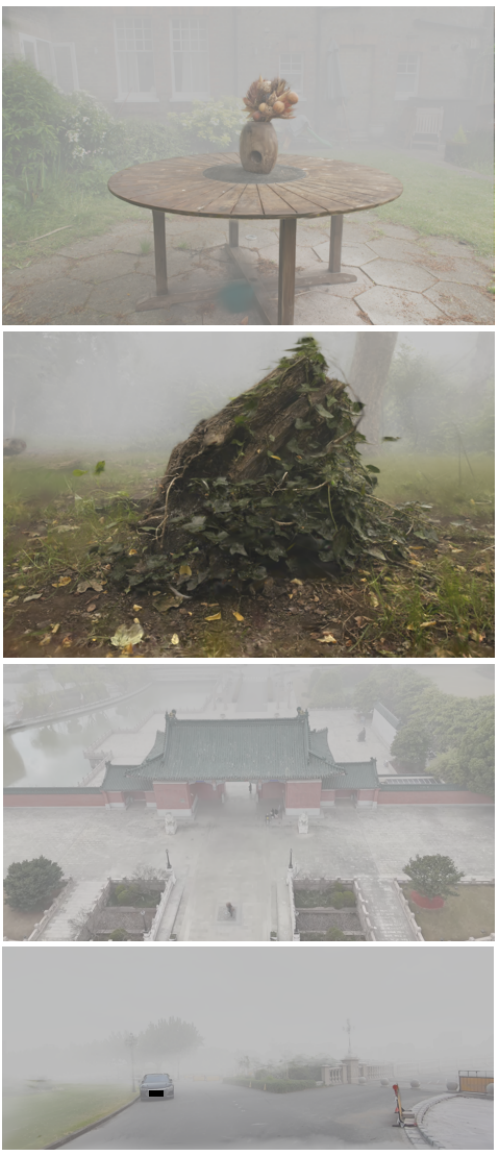}
        \label{fig_dataset_fog}
    }
    \hspace{-3mm}
    \subfloat[Smog]{
        \includegraphics[width=0.14\textwidth]{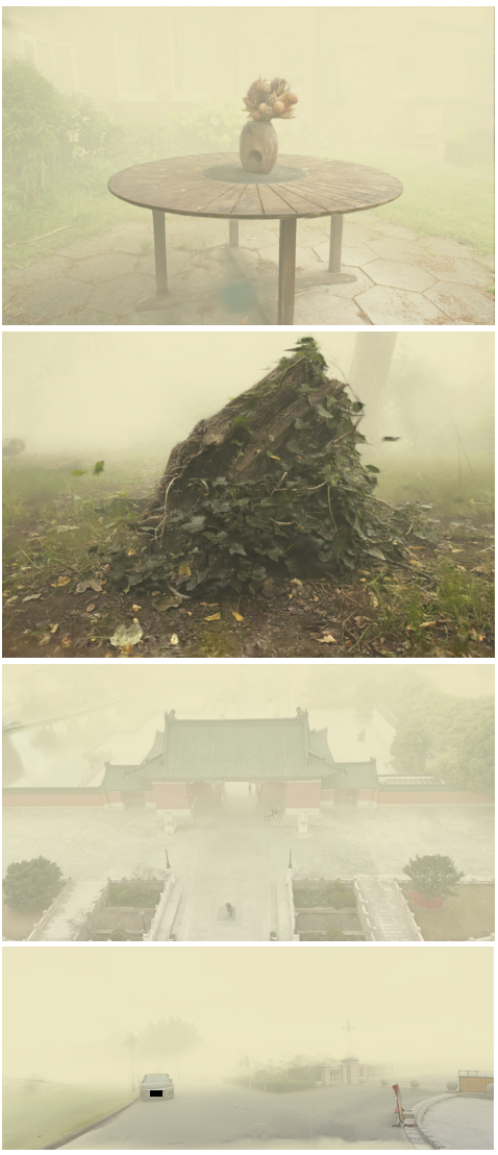}
        \label{fig_dataset_smog}
    }
    \hspace{-3mm}
    \subfloat[Rainfall]{
        \includegraphics[width=0.14\textwidth]{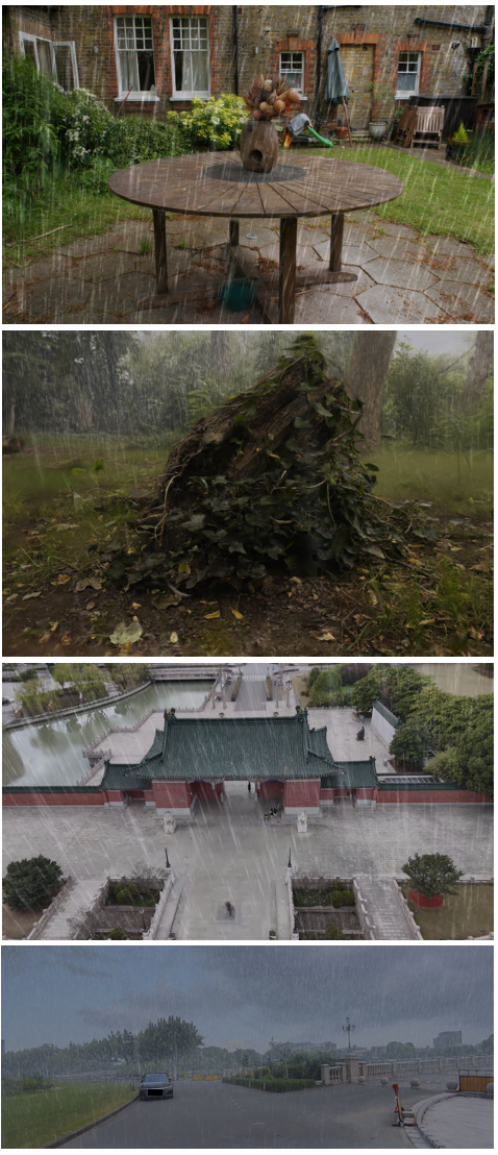}
        \label{fig_dataset_rainfall}
    }
    \hspace{-3mm}
    \subfloat[Snow]{
        \includegraphics[width=0.14\textwidth]{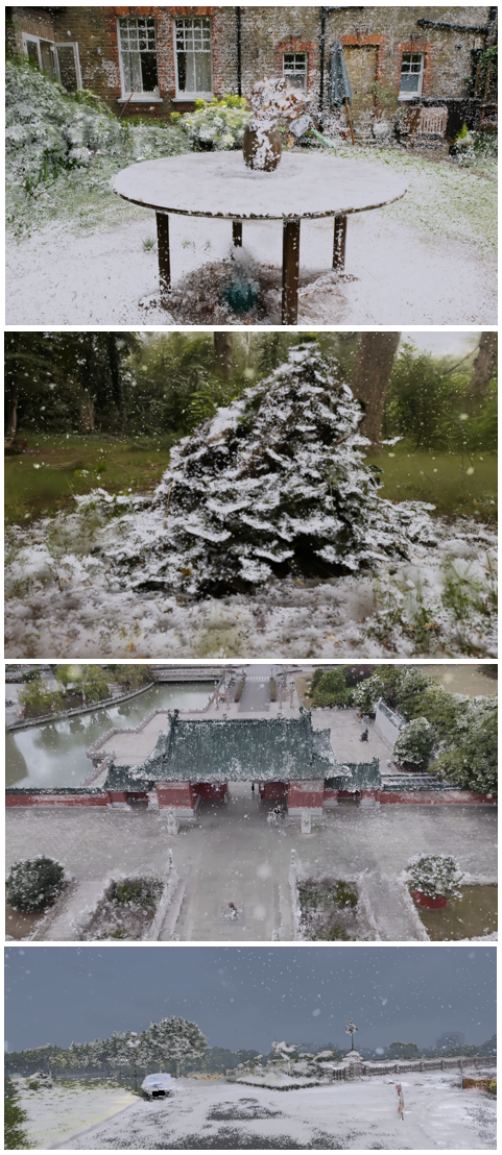}
        \label{fig_dataset_snow}
    }
    \caption{\textbf{Qualitative results on each dataset.} From top to bottom, the scenes in each row are from Mip-NeRF360\cite{barron2022mip} with the \textit{Garden}, \textit{Stump}, SJTU's main gate, and a road driving scene from SJTU campus. In the last column, for the snow effect simulation, we have merged both snowfall and snow accumulation effects to achieve higher realism.}
    \label{fig_dataset}
\end{figure*}

We first tested our weather simulation method on Mip-NeRF360, the traffic and drone scene of SJTU. Fig. \textcolor{red}{\ref{fig_dataset}} shows our render results under different weather types. As the figure represents, our work provides a rich set of weather synthesis options, and is suitable for various outdoor scene scales. It demonstrates particularly robust and excellent rendering results on scenes of small scale. In larger scenes, such as aerial drone footage and driving scenes, it can be observed that effects like fog clearly introduce distant scene blurring. For rain and snow simulations, a simple color filters can make the scene match the desired weather style. Besides the scenes mentioned in the articles, we also provide some extra weather simulation results for several landscapes across the world. The details can be found on our project homepage.


\subsection{\bf Static Weather Synthesis Comparison}

\begin{figure*}[t!]
    \centering
    \subfloat[Original]{
        \includegraphics[height=2in]{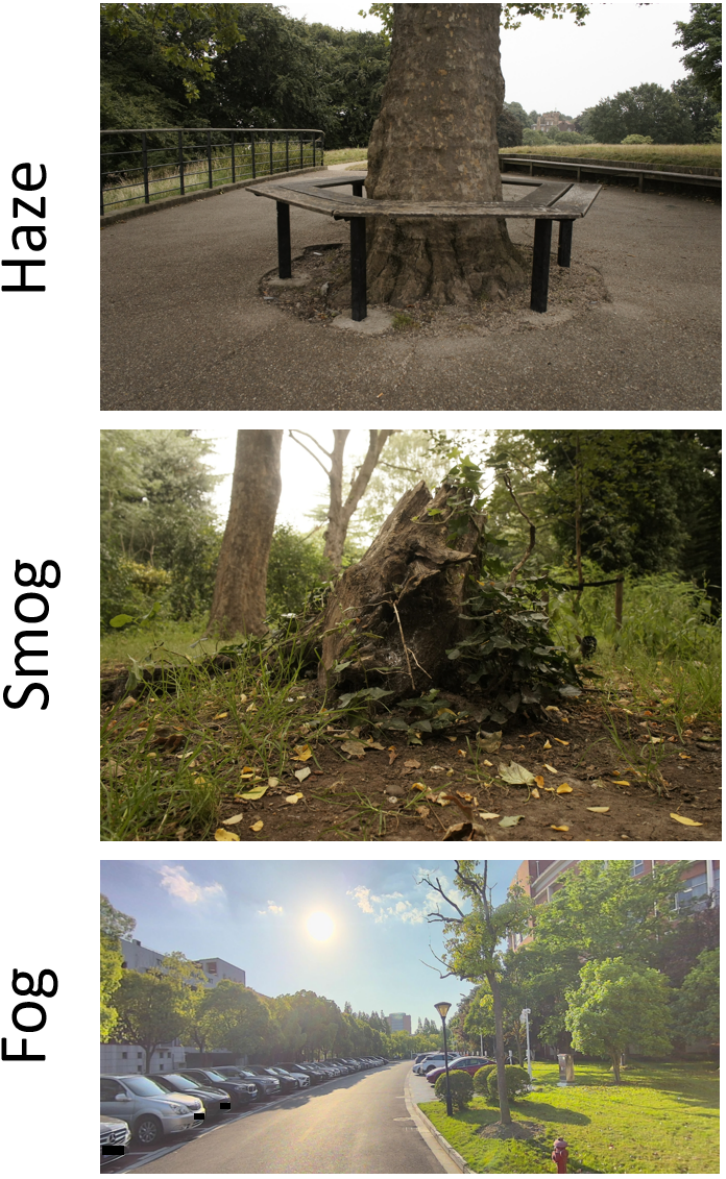}
        \label{fig_fog_clean}
    }
    \hspace{-3mm}
    \subfloat[ClimateGAN\cite{schmidt2021climategan}]{
        \includegraphics[height=2in]{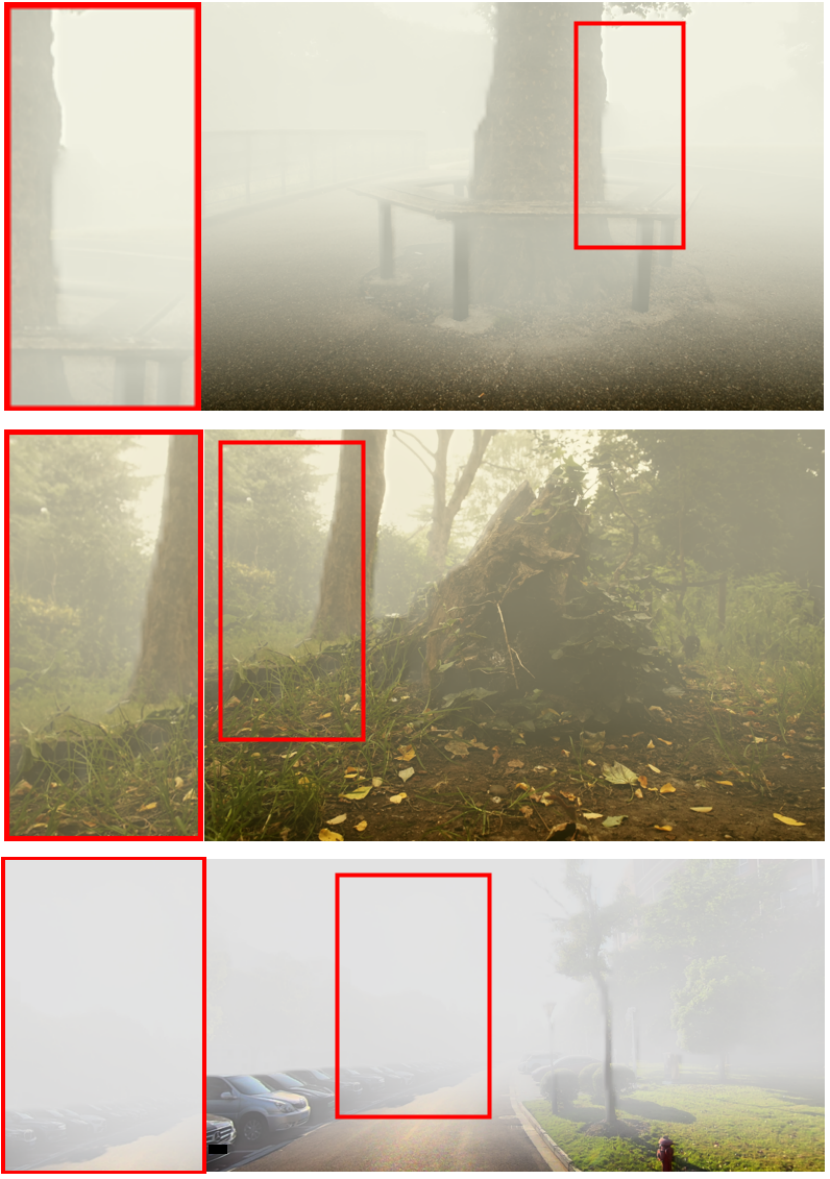}
        \label{fig_fog_gan}
    }
    \hspace{-3mm}
    \subfloat[ClimateNeRF\cite{li2023climatenerf}]{
        \includegraphics[height=2in]{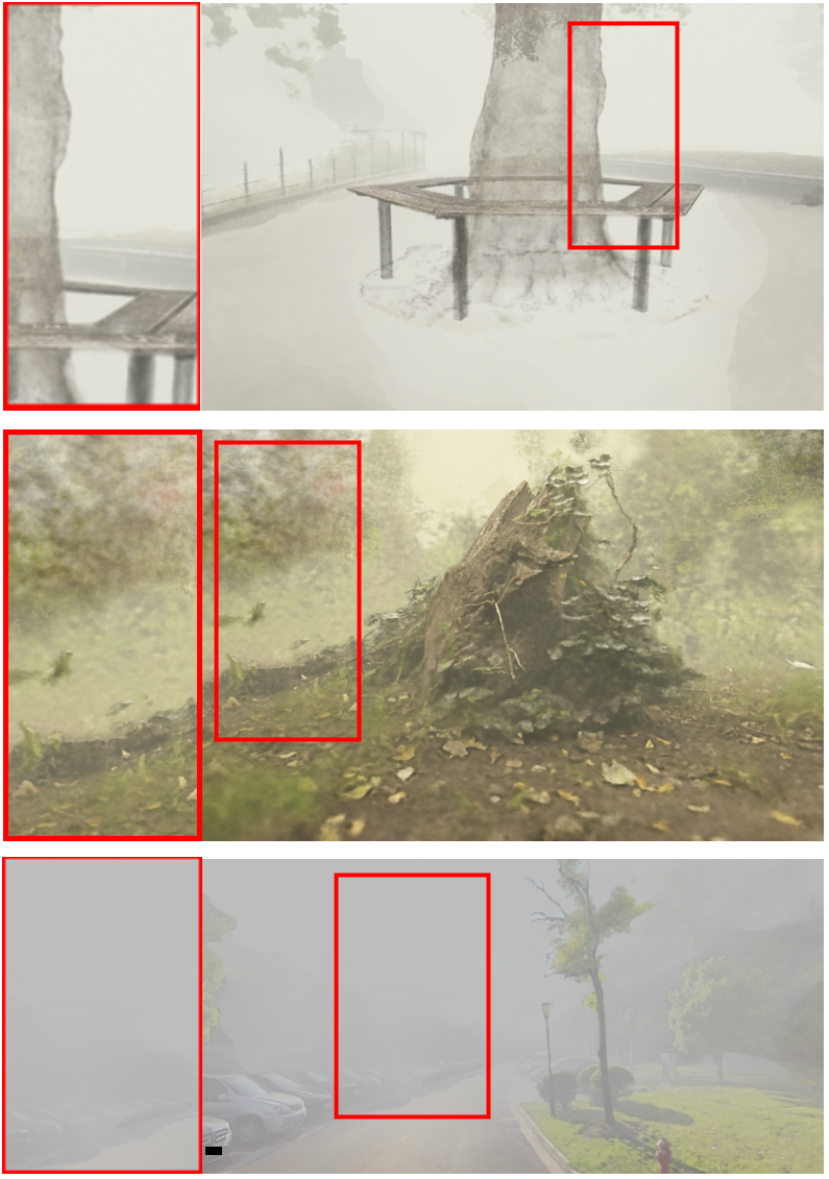}
        \label{fig_fog_nerf}
    }
    \hspace{-3mm}
    \subfloat[Ours]{
        \includegraphics[height=2in]{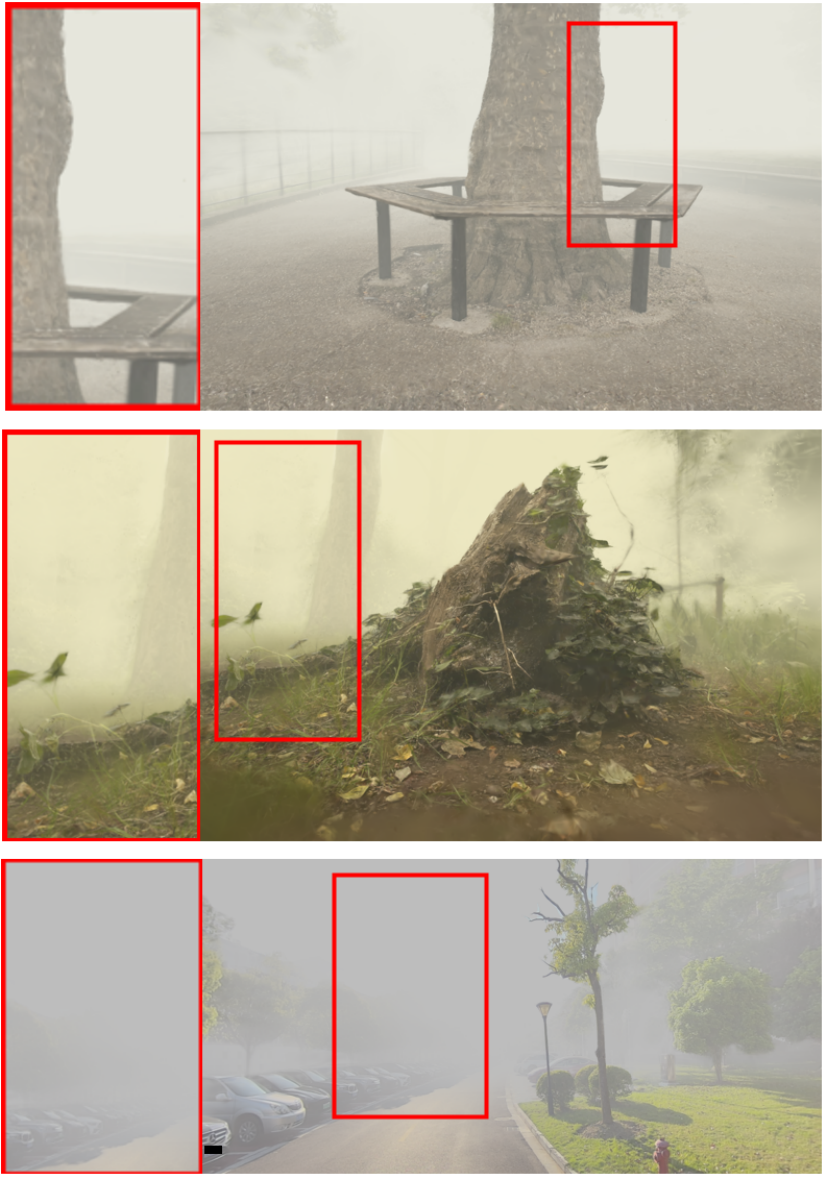}
        \label{fig_fog_ours}
    }
    \caption{\textbf{Static weather simulation comparison.} This figure shows the three blurring effects simulated on three different scenes. ClimateGAN shows inability to deal with depth changes and can be easily distinguished. ClimateNeRF usually shows depth error and can not blur the object in distance completely. Our method shows great depth-based changes, thus enables realistic simualtion.
}
    \label{fig_ex_fog}
\end{figure*}

 This section presents a comparison of static weather effects among different baselines. Notably, ClimateGAN and ClimateNeRF were originally designed to generate smog effects. Here, their color and intensity settings were changed to simulate different blurring effects. As shown in Fig. \textcolor{red}{\ref{fig_ex_fog}}, our method exhibits excellent depth-dependent blurring effects, with a clear and accurate border among objects of different depths. The result of ClimateGAN lacks depth-based changes, with distant scenes appearing as if covered by a grayish mask. ClimateNeRF demonstrates relatively realistic simulation results, but it similarly lacks prior information about the scene depth and frequently shows incorrect depth estimation in certain detailed areas.

\subsection{\bf Dynamic Weather Synthesis Comparison}
\begin{figure}[ht]
\centering
\subfloat[Original]{
    \includegraphics[height=1.36in]{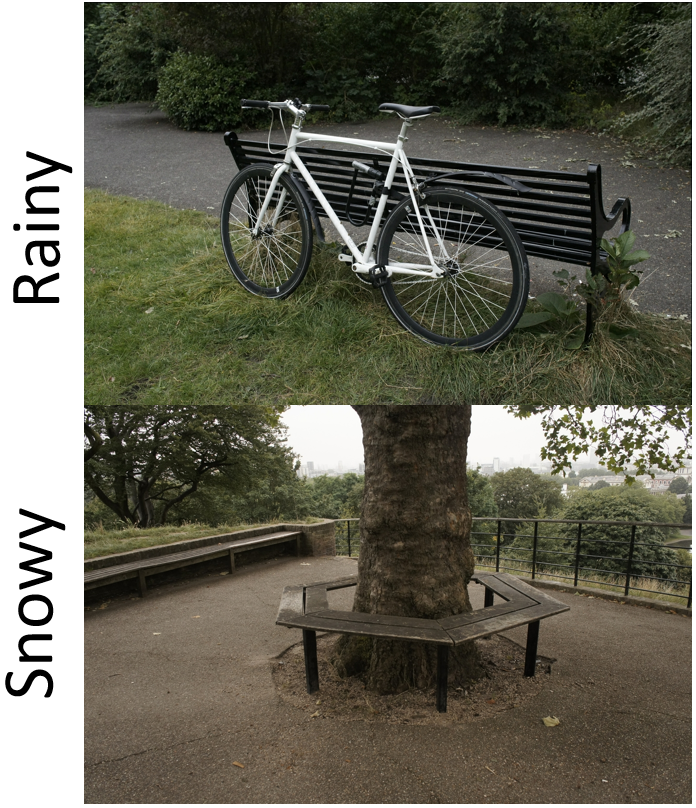}
    \label{fig_rs_clean}
}
\subfloat[SD\cite{rombach2022high}]{
    \includegraphics[height=1.36in]{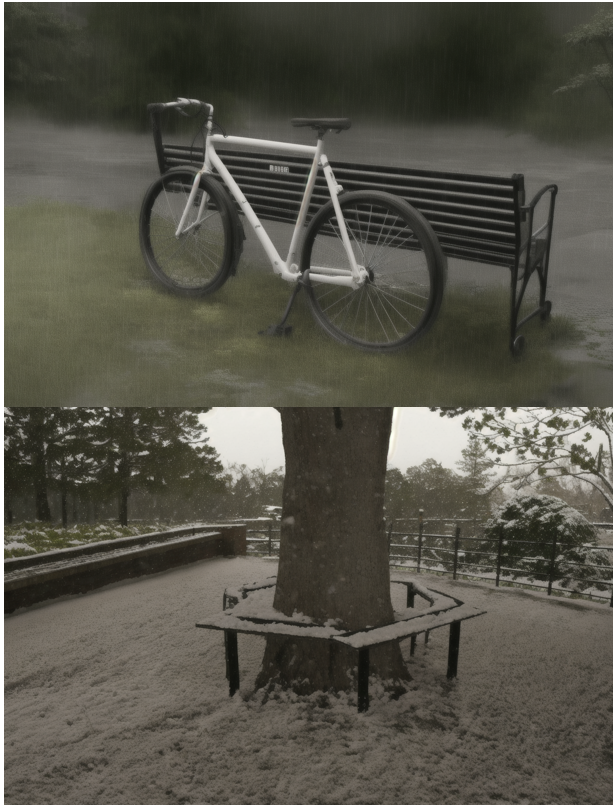}
    \label{fig_rs_sd}
}
\subfloat[Ours]{
    \includegraphics[height=1.36in]{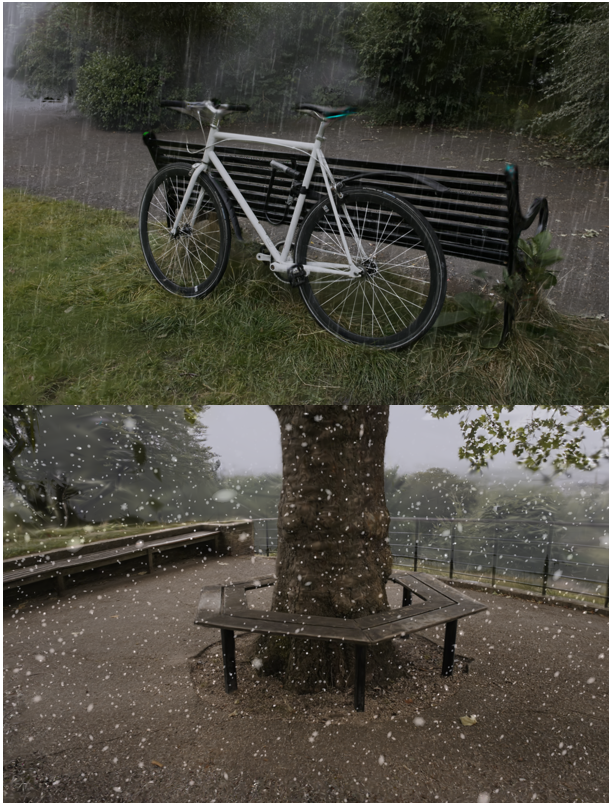}
    \label{fig_rs_ours}
}
\caption{\textbf{Dynamic weather synthesis comparison.} SD may alter original contents and is hard to control the intensity of raindrops or snowflakes. Our method can easily control these effects. 
}
\label{fig_ex_rs}
\end{figure}

For dynamic weather, our method focus on the two most common weather: rainfall and snowfall. For the SD model, the text prompts like "rainstorm" and "snowstorm" were used to control the generated content. 

Fig. \textcolor{red}{\ref{fig_ex_rs}} represents the simulated effects of rainfall and snowfall in the \textit{bicycle} and \textit{treehill} scenes, respectively. While the SD model demonstrates high realism in terms of overall weather style, it struggles to accurately control details such as the size and visibility of each raindrop or snowflake. Additionally, without highly accurate inpainting masks, SD often fails to preserve the real scene contents, and does not provide view-consistent results. In contrast, our method can generate falling elements, with highly customizable parameters such as density and size. Furthermore, our approach supports the generation of dynamic falling effects by applying offsets to each element, making it suitable for video synthesis.

\subsection{\bf Snow Accumulation Simulation Comparison}

\begin{figure*}[t!]
    \centering
    \subfloat[Original]{
        \includegraphics[height=1.7in]{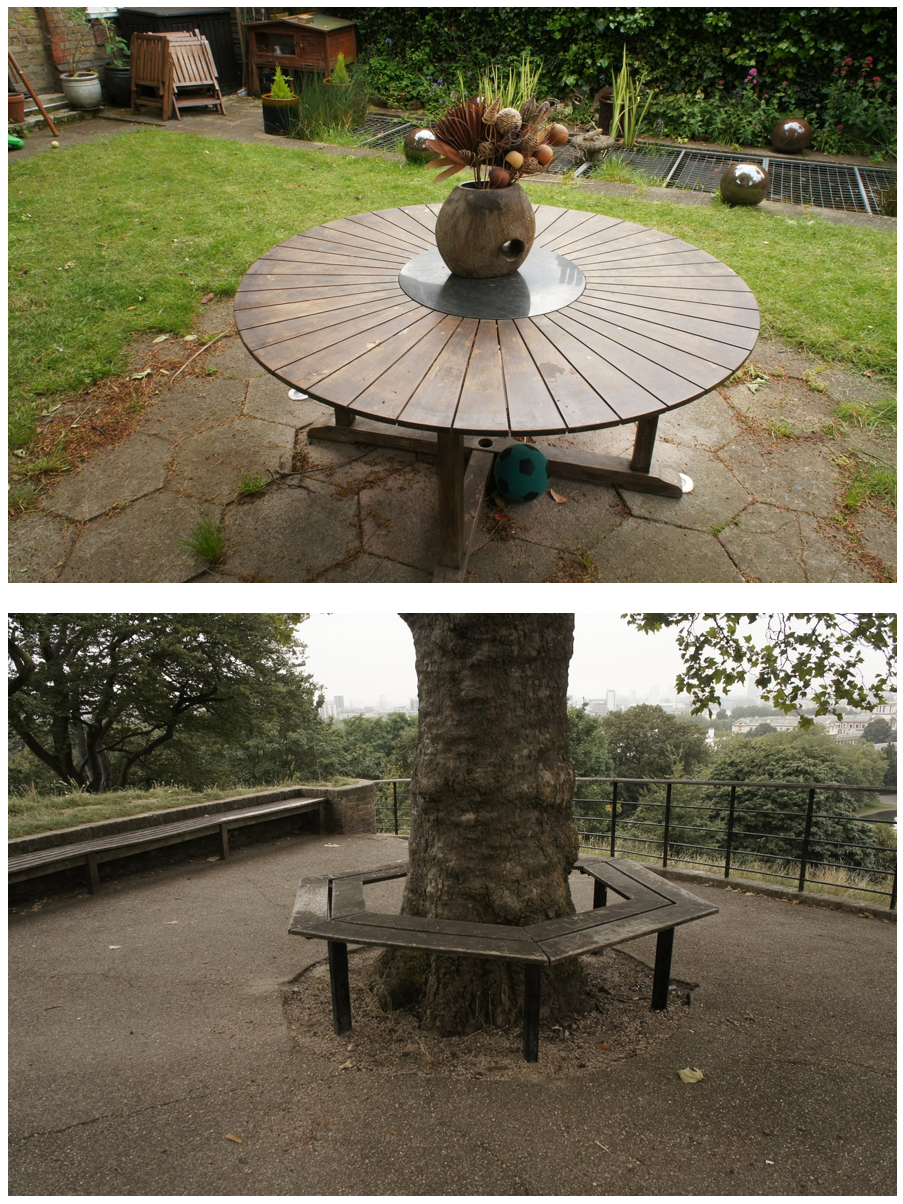}
        \label{fig_ex_sc_clean}
    }
    \hspace{-3mm}
    \subfloat[SD\cite{rombach2022high}]{
        \includegraphics[height=1.7in]{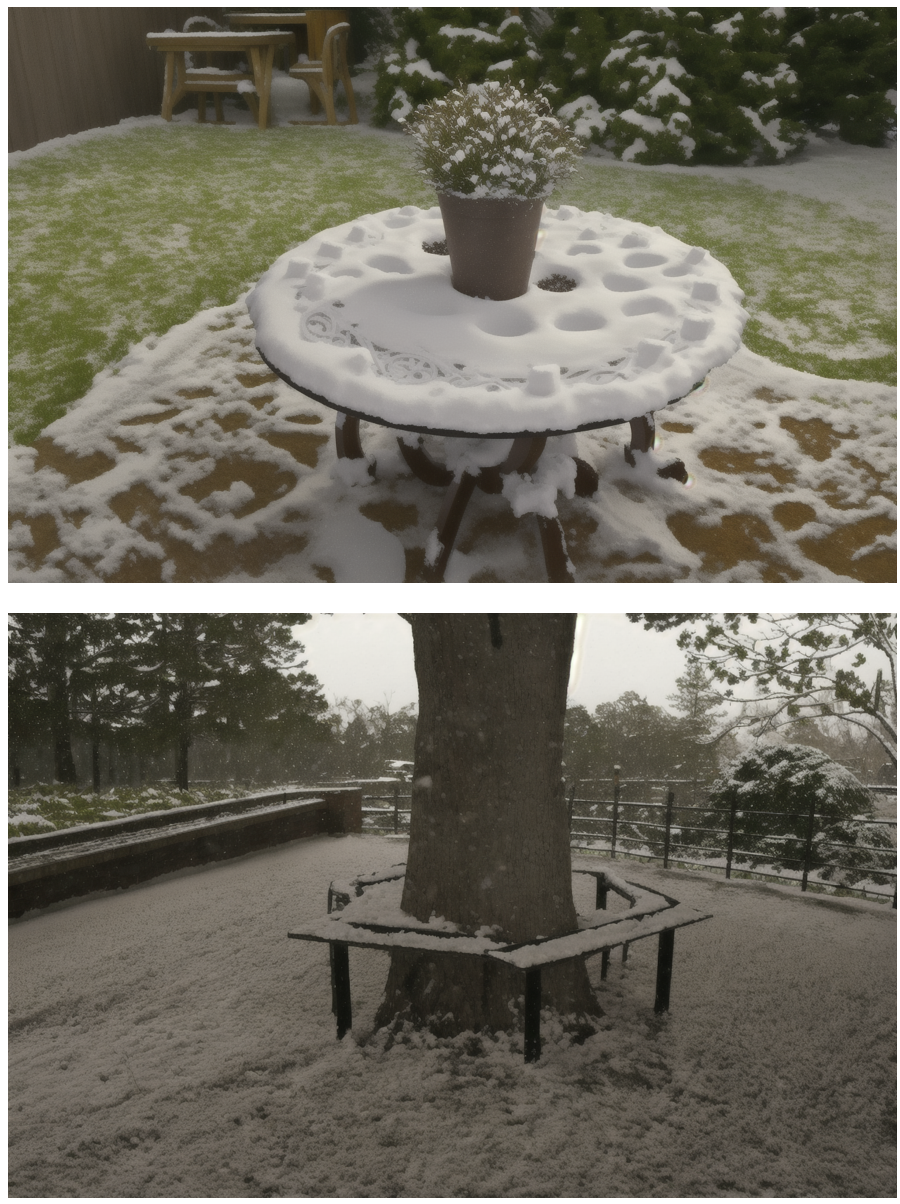}
        \label{fig_ex_sc_sd}
    }
    \hspace{-3mm}
    \subfloat[ClimateNeRF\cite{li2023climatenerf}]{
        \includegraphics[height=1.7in]{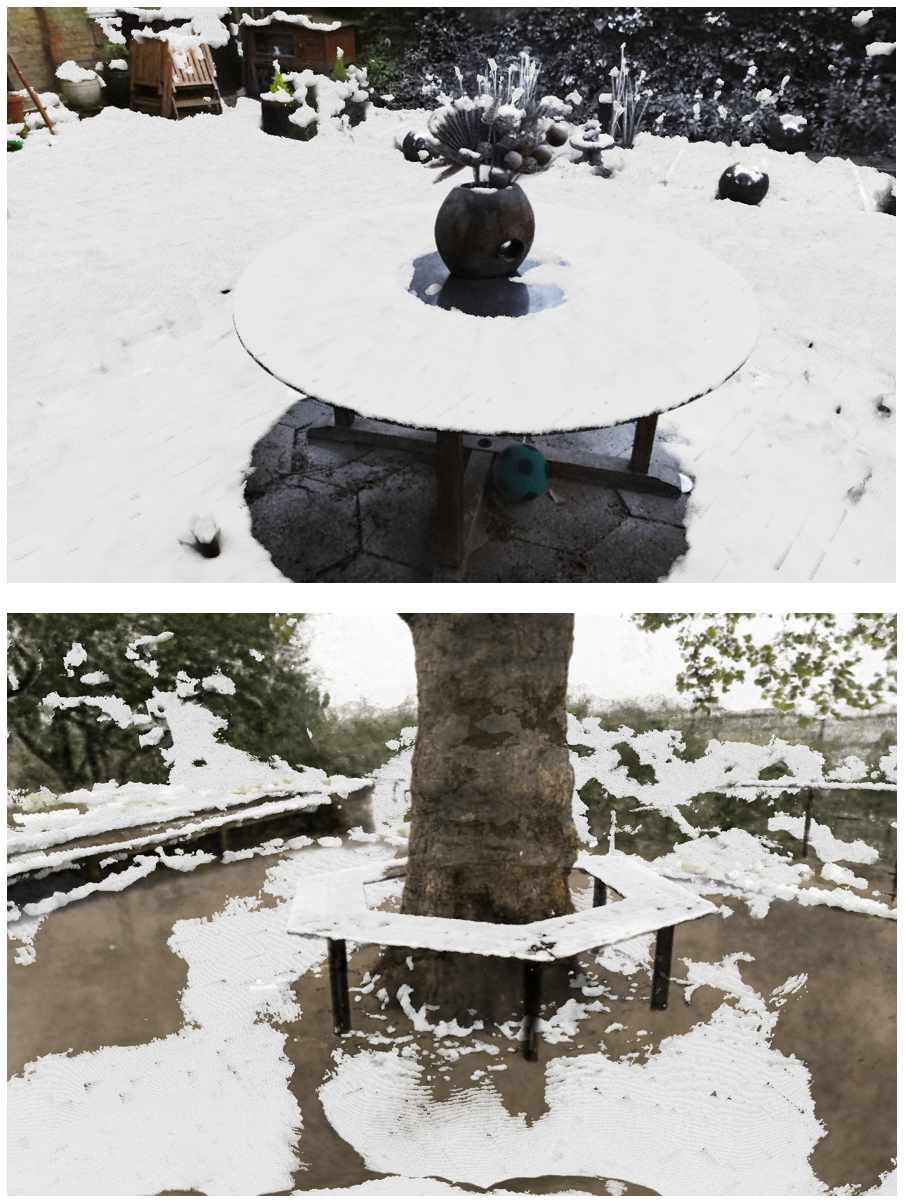}
        \label{fig_ex_sc_nerf}
    }
    \hspace{-3mm}
    \subfloat[Ours]{
        \includegraphics[height=1.7in]{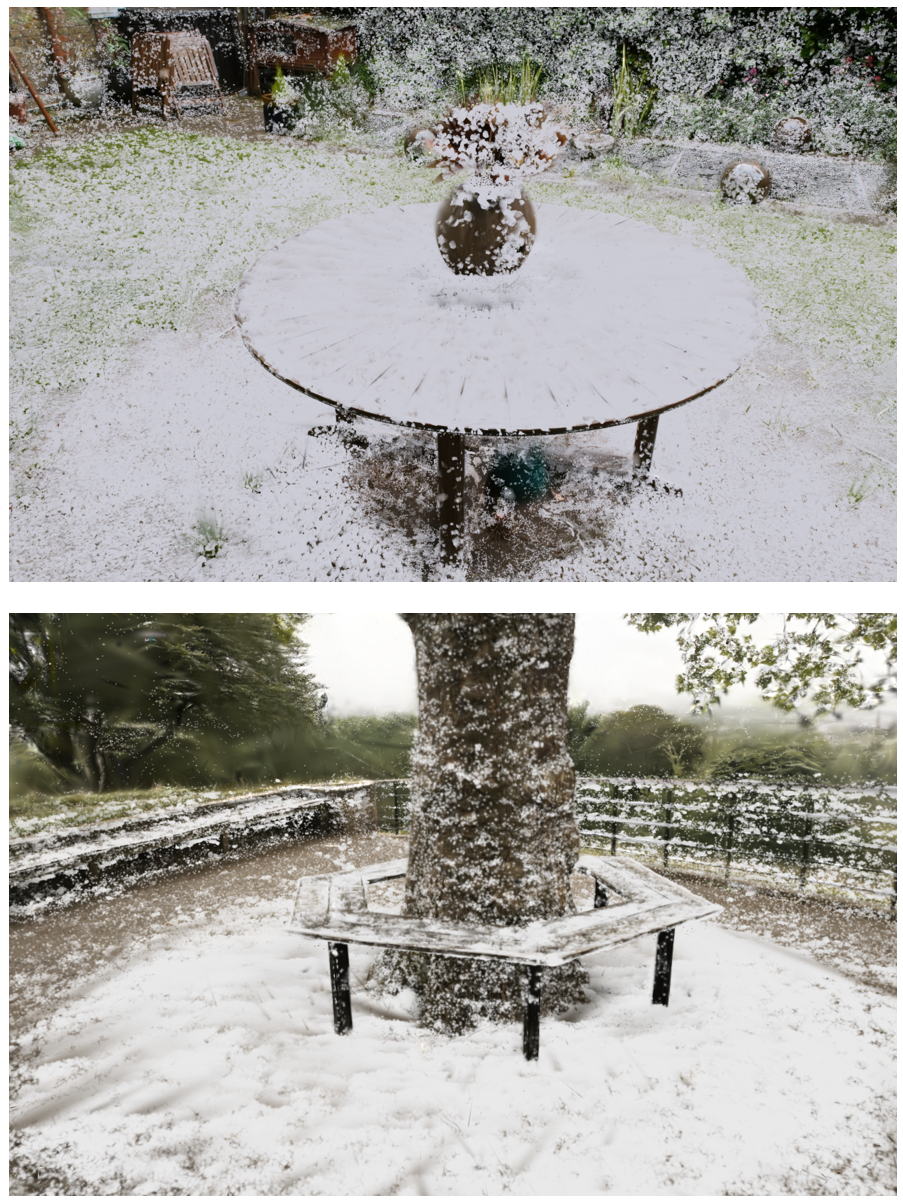}
        \label{fig_ex_sc_ours}
    }
    \caption{\textbf{Snow accumulation simulation comparison.} The snow results from SD exhibited significant alterations to the scene objects. Both ClimateNeRF and our method preserved the original scene characteristics, but in some cases, incorrect depth estimation in ClimateNeRF led to erroneous snow effects.
    }
    \label{fig_ex_sc}
\end{figure*}

The comparative results of snow accumulation simulations are presented in Fig. \textcolor{red}{\ref{fig_ex_sc}}.  As the result shows, SD exhibit significant instability, often altering scene objects and is hard to control the density of the snow cover. ClimateNeRF can model scenes and simulate high-quality snow-covered surfaces with a high degree of smoothness. However, this comes at the cost of substantial speed loss, which will be quantitatively analyzed later. Additionally, as shown in the lower part of Fig. \textcolor{red}{\ref{fig_ex_sc_nerf}}, for the \textit{treehill} scene, ClimateNeRF's incorrect depth estimation results in chaotic simulation results.

In contrast, our method delivers robust and complete snow accumulation effects across typical scenes, with most snow points correctly laid on geometric surfaces. Furthermore, even with a substantial number of snow Gaussians, our method ensures real-time performance, which will be demonstrated later.

\subsection{\bf View Consistency Comparison}

\begin{figure}[ht]
    \centering
    \includegraphics[width=0.36\textwidth]{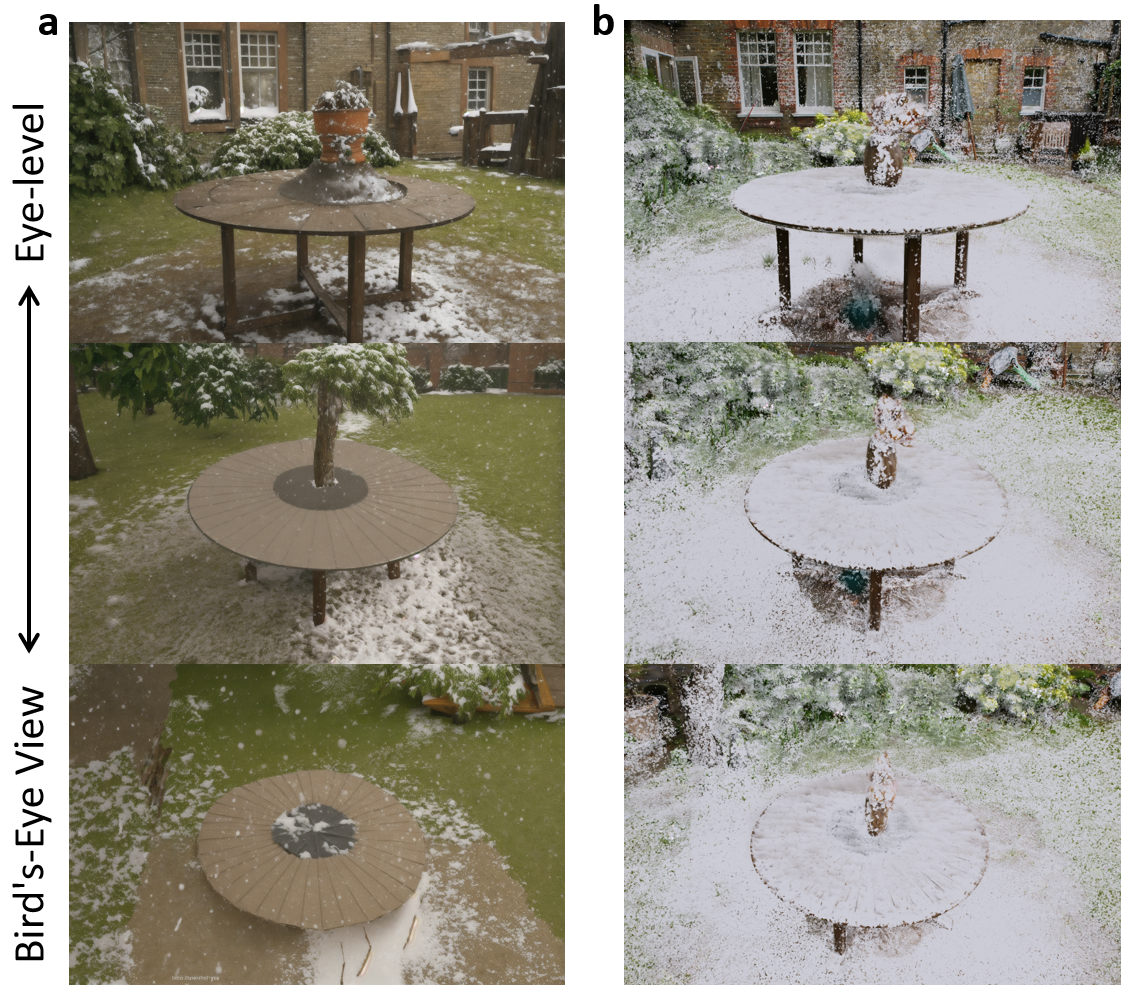}
    \caption{\textbf{Weather simulation images from different viewpoints.} \textbf{a} shows images of the \textit{garden} scene generated by SD\cite{rombach2022high} from different viewpoints. \textbf{b} represents images simulated and rendered using our method. 
    }
    \label{fig_ex_consistency}
\end{figure}

We validate the viewpoint consistency of our method through both qualitative experiments. As shown in Fig. \textcolor{red}{\ref{fig_ex_consistency}}, the table and flower pot generated by SD always change their original shapes and appearances. Single-frame generation methods often fail to ensure view-consistency across different viewpoint inputs, making them unsuitable for tasks such as video synthesis. In contrast, our approach effectively preserves the background content while preserving high rendering quality.

\subsection{\bf Efficiency Comparison}

\begin{table*}[!t]
\caption{Rendering/Generation Speed (FPS)($\uparrow$)\label{tab:speed}}
\centering
\setlength{\arrayrulewidth}{1pt}
\arrayrulecolor{black}
\begin{tabular}{c c c c c}
\hline
\textbf{Method} & \textbf{Clear weather} & \textbf{Fog / Smog / Haze simulation} & \textbf{Rain / Snowfall simulation} & \textbf{Snow cover simulation}\\ 
\hline
ClimateGAN\cite{schmidt2021climategan} & $\backslash$ & 0.11 & $\backslash$ & $\backslash$ \\
SD\cite{rombach2022high} & 0.12 & 0.12 & 0.12 & 0.12 \\
ClimateNeRF\cite{li2023climatenerf} & 2.26 & 1.33 & $\backslash$ & 0.09 \\
Ours & \textbf{83.30} & \textbf{31.27} & \textbf{10.42} & \textbf{58.24}\\
\hline
\end{tabular}
\end{table*}

Table \textcolor{red}{\ref{tab:speed}} presents the rendering (generation) speed per frame during the weather editing tasks on the Mip-NeRF360 scenes with RTX 4090. For the experiments, we utilized the pretrained weights provided by ClimateGAN and pretrained generalized model of SD, while the sampling steps for SD were set to 20, with the output images maintaining the same dimensions as the input. We have applied the img2img method for SD experiments. Additionally, both methods had their batch sizes set to 1 to ensure a fair comparison under controlled conditions. The output image size is set to $1200\times800$.

As shown, our method demonstrates real-time rendering performance with different simulated weather effects. In contrast, the generation speeds of ClimateGAN and SD are significantly slower, with a speed gap of up to 100 times compared to our method. The rendering speed of ClimateNeRF is limited by multiple sampling steps, making real-time rendering very challenging. This becomes more obvious when handling snow accumulation, where hours are needed to synthesize a short video. In scenarios requiring repetitive tests, the real-time performance of our method provides a significant advantage. It is worth noting that our tests on rainfall and snowfall show performance degradation due to the need to render multiple noise scenes within a single frame. In order to regain the performance, it is possible to decrease the number of noise scenes or merge noise scenes and the original scene, which may generate some color or occlusion errors because of the opacity issues mentioned in the methods parts.

\subsection{\bf Quantitative Evaluation of Simulation Quality}

\begin{table*}[!t]
\caption{CMMD scores comparison ($\downarrow$)\label{tab:cmmd}}
\centering
\setlength{\arrayrulewidth}{1pt}
\arrayrulecolor{black}
\begin{tabular}{c c c c}
\hline
\textbf{Method} & \textbf{Fog / Haze / Smog simulation} & \textbf{Rainfall simulation} & \textbf{Snow simulation} \\
\hline
ClimateGAN\cite{schmidt2021climategan}  & 1.84 & $\backslash$ & $\backslash$ \\
DID-MDN\cite{zhang2018density}          & $\backslash$ & 3.14 & $\backslash$ \\
SD\cite{rombach2022high}                & 2.32 & 3.12 & \textbf{2.66} \\
ClimateNeRF\cite{li2023climatenerf}     & 2.11 & $\backslash$  & 2.96 \\
Ours                       & \textbf{1.40} & \textbf{2.79} & 3.30 \\
\hline
\end{tabular}
\end{table*}


Quantitatively evaluating the quality of weather simulation is quite challenging, as there is currently no well-established metric to evaluate the realism of weather. For reference purposes, we have chosen the CMMD metric\cite{jayasumana2024rethinking}, which quantitatively measures the distance between the data distributions. We conducted the experiment on the four real weather dataset mentioned in the part IV.A. Among the comparison methods, we included DID-MDN\cite{zhang2018density}, a synthetic rain dataset as benchmark with a total of 1,800 frames. Table \textcolor{red}{\ref{tab:cmmd}} shows different CMMDs, which are computed between the render results of different methods and the real datasets. Here, the snowfall effect and snow accumulation are merged into one category because real snowy datasets always include both phenomena. We did not include weather images generated from pure noise by SD but applied SD inpainting on the original scene images for evaluation, which ensures a fair comparison. 

It can be seen that our method demonstrate lower CMMD on both the static and dynamic weather simulation results. This result indicates that our simulated rendering outcomes could potentially be applied in areas such as image data augmentation, which remains to be further experimentally validated. The relatively higher CMMD for our snow cover simulation may partly result from the overall stylistic and tonal differences between the simulated scene and real snowy scenes.

\begin{table*}[!t]
\caption{UVQ MOS comparison ($\uparrow$)\label{tab:uvq}}
\centering
\setlength{\arrayrulewidth}{1pt}
\arrayrulecolor{black}
\begin{tabular}{c | c  c  c  c | c  c | c  c  c}
\hline
\textbf{} & \multicolumn{4}{c |}{\textbf{Fog / Haze / Smog simulation}} & \multicolumn{2}{c |}{\textbf{Rainfall simulation}} & \multicolumn{3}{c}{\textbf{Snow simulation}} \\
\hline
\textbf{Scene} & SD\cite{rombach2022high}  & ClimateGAN\cite{schmidt2021climategan} & ClimateNeRF\cite{li2023climatenerf} & Ours & SD & Ours & SD & ClimateNeRF & Ours \\
\hline
\textit{Stump}             & 3.41 & 3.56 & 3.23 & \textbf{4.43} & 3.60 & \textbf{3.81} & 3.47 & 3.62 & \textbf{3.79} \\
\textit{Bicycle}           & 3.97 & 4.12 & 4.10 & \textbf{4.23} & 3.96 & \textbf{4.20} & 3.81 & \textbf{3.94} & 3.79 \\
\textit{Garden}             & 3.75 & 4.10 & \textbf{4.17} & 3.87 & 3.77 & \textbf{3.97} & 3.66 & \textbf{3.85} & 3.74 \\
\textit{Treehill}           & 3.43 & 3.65 & 3.41 & \textbf{3.96} & 3.33 & \textbf{3.75} & 3.32 & 3.60 & \textbf{3.62} \\
\textit{SJTU road driving}  & 3.76 & 3.85 & 3.74 & \textbf{4.12} & 3.72 & \textbf{4.21} & 3.74 & 4.41 & \textbf{4.49} \\
\textit{SJTU main gate}     & 3.45 & \textbf{4.14} & 3.37 & 3.96 & 3.29 & \textbf{3.82} & 3.63 & 3.91 & \textbf{3.95} \\
\hline
\end{tabular}
\end{table*}


We also used the UVQ (Universal Video Quality) model\cite{wang2021rich} to evaluate the quality of the videos synthesized by different methods, using the “compression content distortion” score from its predictions as the default score for the Mean Opinion Score (MOS). From worst to best, the score ranges from 1 to 5. The evaluation results are presented in Table \textcolor{red}{\ref{tab:uvq}}. We have evaluated the video quality for fog, rainfall, and snow accumulation conditions. We have not separately evaluated the video quality for snowfall and snow cover, as SD often merges those as one type. 

It can be seen that our method achieved better scores in most of the results. Although in certain scenes, the qualitative results showed that our method performed better than others, it still did not achieve better scores in the table (e.g., snow in \textit{Treehill}). Also, while SD actually does not preserve the consistency across frames, its videos show relatively good scores. Those suggest that UVQ may not yet be a representative metric for evaluating weather simulation quality and the video consistency. A more accurate evaluation criteria for weather simulation needs to be further studied.

\subsection{\bf Controllability}

\begin{figure}[ht]
    \centering
    \includegraphics[width=0.4\textwidth]{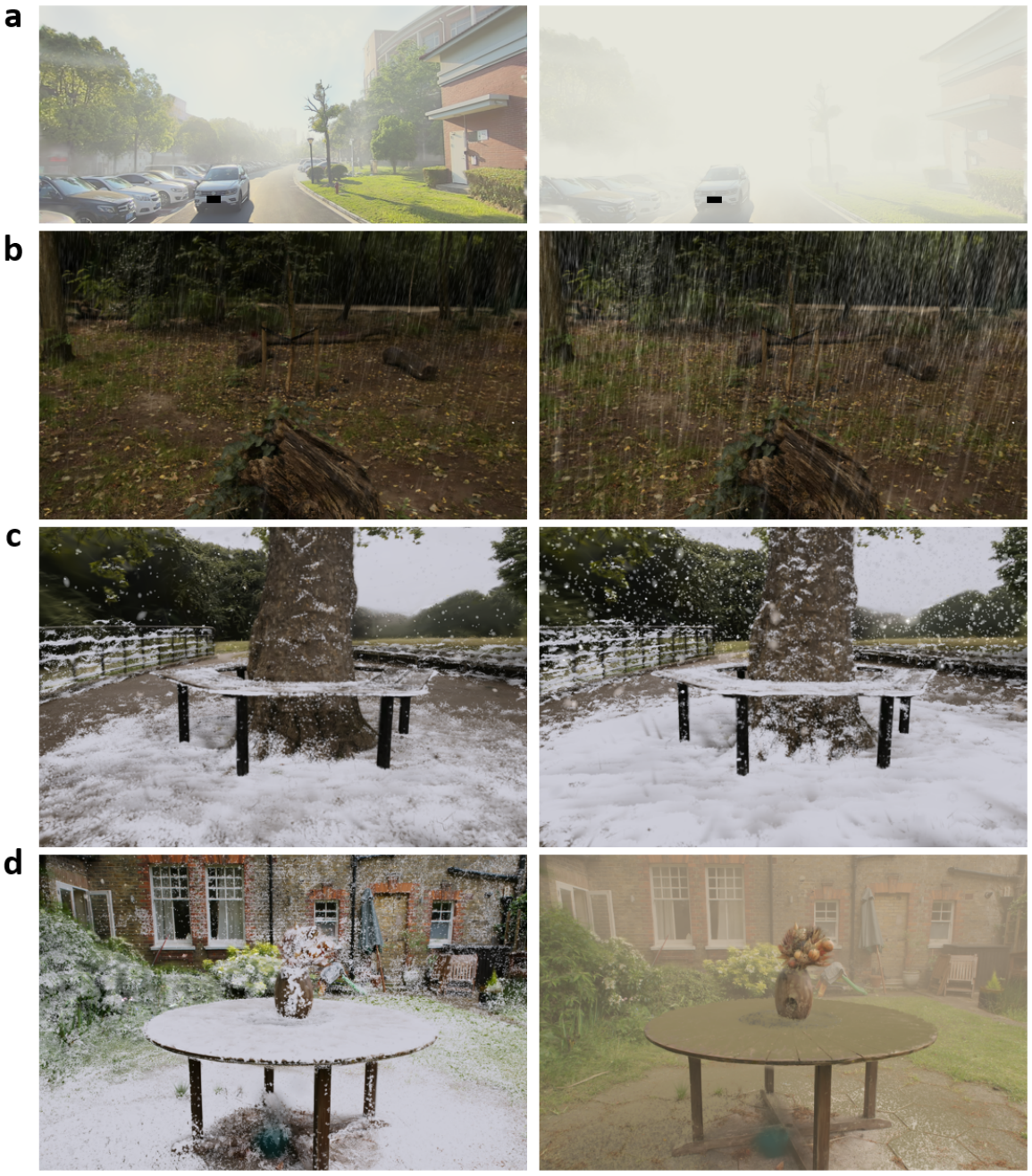}
    \label{fig_ex_control_inside}
    \caption{\textbf{Controllability. } \textbf{a} proves the possibility to modify the intensity when applying blurring effect. \textbf{b} shows the different transparency of noise particles(here raindrops). \textbf{c} and \textbf{d} represent the different snowfall intensity, snow cover density and the case where we change snow color to simulate dust accumulation in sandstorm weather.
}
    \label{fig_ex_control}
\end{figure}

Regarding the controllability of our method, Fig. \textcolor{red}{\ref{fig_ex_control}} demonstrates results under various control settings. By adjusting the parameters before rendering, our approach can flexibly achieve various weather variations, such as detailed changes in the overall weather intensity and colors. In video synthesis, it is possible to adjust the direction and speed of rainfall/snowfall, allowing for flexible simulation under different scenarios.

\section{Conclusion and Discussion}
In this work, we proposed a framework based on 3DGS for synthesize various common weather effects. By leveraging render results of the optimized Gaussian scene, combined with post-processing, Gaussian insertion, and editing operations, we successfully achieved the simulation and editing of weather such as fog, haze, smog, rainfall, snowfall, and snow accumulation. Our method effectively generates highly realistic weather images while ensuring view-consistency, enabling the synthesis of coherent and high-quality videos. With its real-time rendering speed and highly customizable weather parameters, it allows users to flexibly test and adjust weather styles. In addition, the high-fidelity simulated data synthesized by our work can work as challenging corner cases for perception tasks in open scenes under various common weather conditions, such as for the datasets used by autonomous driving\cite{kejriwal2024challenges}.

However, our weather simulation quality partly relies on the based scene reconstruction quality. 3DGS may experience performance degradation in scenarios with dynamic objects or large-scale complex scenes, leading to a range of issues, such as uneven blurring intensity and incorrect occlusion. In the future, we will continue our research to improve the stability of reconstruction results across various scenes and to achieve more robust weather simulation effects.

\bibliographystyle{IEEEtran}

\begin{thebibliography}{10}
\bibitem{blender}
B.~O. Community, \emph{Blender - a 3D modelling and rendering package}, Blender Foundation, Stichting Blender Foundation, Amsterdam, 2018. [Online]. Available: \url{http://www.blender.org}

\bibitem{dosovitskiy2017carla}
A.~Dosovitskiy, G.~Ros, F.~Codevilla, A.~Lopez, and V.~Koltun, ``Carla: An open urban driving simulator,'' in \emph{Conference on robot learning}.\hskip 1em plus 0.5em minus 0.4em\relax PMLR, 2017, pp. 1--16.

\bibitem{shah2018airsim}
S.~Shah, D.~Dey, C.~Lovett, and A.~Kapoor, ``Airsim: High-fidelity visual and physical simulation for autonomous vehicles,'' in \emph{Field and Service Robotics: Results of the 11th International Conference}.\hskip 1em plus 0.5em minus 0.4em\relax Springer, 2018, pp. 621--635.

\bibitem{noueihed2022knowledge}
H.~Noueihed, H.~Harb, and J.~Tekli, ``Knowledge-based virtual outdoor weather event simulator using unity 3d,'' \emph{The Journal of Supercomputing}, vol.~78, no.~8, pp. 10\,620--10\,655, 2022.

\bibitem{engine2018unreal}
U.~Engine, ``Unreal engine,'' \emph{Retrieved from Unreal Engine: https://www. unrealengine. com/en-US/what-is-unreal-engine-4}, 2018.

\bibitem{schonberger2016structure}
J.~L. Schonberger and J.-M. Frahm, ``Structure-from-motion revisited,'' in \emph{Proceedings of the IEEE conference on computer vision and pattern recognition}, 2016, pp. 4104--4113.

\bibitem{goesele2006multi}
M.~Goesele, B.~Curless, and S.~M. Seitz, ``Multi-view stereo revisited,'' in \emph{2006 IEEE Computer Society Conference on Computer Vision and Pattern Recognition (CVPR'06)}, vol.~2.\hskip 1em plus 0.5em minus 0.4em\relax IEEE, 2006, pp. 2402--2409.

\bibitem{mildenhall2021nerf}
B.~Mildenhall, P.~P. Srinivasan, M.~Tancik, J.~T. Barron, R.~Ramamoorthi, and R.~Ng, ``Nerf: Representing scenes as neural radiance fields for view synthesis,'' \emph{Communications of the ACM}, vol.~65, no.~1, pp. 99--106, 2021.

\bibitem{kerbl20233d}
B.~Kerbl, G.~Kopanas, T.~Leimk{\"u}hler, and G.~Drettakis, ``3d gaussian splatting for real-time radiance field rendering.'' \emph{ACM Trans. Graph.}, vol.~42, no.~4, pp. 139--1, 2023.

\bibitem{wu20244d}
G.~Wu, T.~Yi, J.~Fang, L.~Xie, X.~Zhang, W.~Wei, W.~Liu, Q.~Tian, and X.~Wang, ``4d gaussian splatting for real-time dynamic scene rendering,'' in \emph{Proceedings of the IEEE/CVF Conference on Computer Vision and Pattern Recognition}, 2024, pp. 20\,310--20\,320.

\bibitem{qian2024weathergs}
C.~Qian, Y.~Guo, W.~Li, and G.~Markkula, ``Weathergs: 3d scene reconstruction in adverse weather conditions via gaussian splatting,'' \emph{arXiv preprint arXiv:2412.18862}, 2024.

\bibitem{liu2024deraings}
S.~Liu, X.~Chen, H.~Chen, Q.~Xu, and M.~Li, ``Deraings: Gaussian splatting for enhanced scene reconstruction in rainy environments,'' \emph{arXiv preprint arXiv:2408.11540}, 2024.

\bibitem{wang2024gaussianeditor}
J.~Wang, J.~Fang, X.~Zhang, L.~Xie, and Q.~Tian, ``Gaussianeditor: Editing 3d gaussians delicately with text instructions,'' in \emph{Proceedings of the IEEE/CVF Conference on Computer Vision and Pattern Recognition}, 2024, pp. 20\,902--20\,911.

\bibitem{wang2025view}
Y.~Wang, X.~Yi, Z.~Wu, N.~Zhao, L.~Chen, and H.~Zhang, ``View-consistent 3d editing with gaussian splatting,'' in \emph{European Conference on Computer Vision}.\hskip 1em plus 0.5em minus 0.4em\relax Springer, 2025, pp. 404--420.

\bibitem{xu2025texture}
T.-X. Xu, W.~Hu, Y.-K. Lai, Y.~Shan, and S.-H. Zhang, ``Texture-gs: Disentangling the geometry and texture for 3d gaussian splatting editing,'' in \emph{European Conference on Computer Vision}.\hskip 1em plus 0.5em minus 0.4em\relax Springer, 2025, pp. 37--53.

\bibitem{rong2024gstex}
V.~Rong, J.~Chen, S.~Bahmani, K.~N. Kutulakos, and D.~B. Lindell, ``Gstex: Per-primitive texturing of 2d gaussian splatting for decoupled appearance and geometry modeling,'' \emph{arXiv preprint arXiv:2409.12954}, 2024.

\bibitem{chen2024dge}
M.~Chen, I.~Laina, and A.~Vedaldi, ``Dge: Direct gaussian 3d editing by consistent multi-view editing,'' in \emph{European Conference on Computer Vision}.\hskip 1em plus 0.5em minus 0.4em\relax Springer, 2024, pp. 74--92.

\bibitem{dai2025rainygs}
Q.~Dai, X.~Ni, Q.~Shen, W.~Chen, B.~Chen, and M.~Chu, ``Rainygs: Efficient rain synthesis with physically-based gaussian splatting,'' \emph{arXiv preprint arXiv:2503.21442}, 2025.

\bibitem{fiebelman2025let}
G.~Fiebelman, H.~Averbuch-Elor, and S.~Benaim, ``Let it snow! animating static gaussian scenes with dynamic weather effects,'' \emph{arXiv preprint arXiv:2504.05296}, 2025.

\bibitem{wu2024gaussctrl}
J.~Wu, J.-W. Bian, X.~Li, G.~Wang, I.~Reid, P.~Torr, and V.~A. Prisacariu, ``Gaussctrl: Multi-view consistent text-driven 3d gaussian splatting editing,'' in \emph{European Conference on Computer Vision}.\hskip 1em plus 0.5em minus 0.4em\relax Springer, 2024, pp. 55--71.

\bibitem{ho2020denoising}
J.~Ho, A.~Jain, and P.~Abbeel, ``Denoising diffusion probabilistic models,'' \emph{Advances in neural information processing systems}, vol.~33, pp. 6840--6851, 2020.

\bibitem{von2019simulating}
A.~Von~Bernuth, G.~Volk, and O.~Bringmann, ``Simulating photo-realistic snow and fog on existing images for enhanced cnn training and evaluation,'' in \emph{2019 IEEE Intelligent Transportation Systems Conference (ITSC)}.\hskip 1em plus 0.5em minus 0.4em\relax IEEE, 2019, pp. 41--46.

\bibitem{nikolov2024digiweather}
I.~Nikolov, ``Digiweather: Synthetic rain, snow and fog dataset augmentation,'' in \emph{International Conference on Extended Reality}.\hskip 1em plus 0.5em minus 0.4em\relax Springer, 2024, pp. 22--41.

\bibitem{goodfellow2020generative}
I.~Goodfellow, J.~Pouget-Abadie, M.~Mirza, B.~Xu, D.~Warde-Farley, S.~Ozair, A.~Courville, and Y.~Bengio, ``Generative adversarial networks,'' \emph{Communications of the ACM}, vol.~63, no.~11, pp. 139--144, 2020.

\bibitem{chen2017stylebank}
D.~Chen, L.~Yuan, J.~Liao, N.~Yu, and G.~Hua, ``Stylebank: An explicit representation for neural image style transfer,'' in \emph{Proceedings of the IEEE conference on computer vision and pattern recognition}, 2017, pp. 1897--1906.

\bibitem{zhang2017style}
L.~Zhang, Y.~Ji, X.~Lin, and C.~Liu, ``Style transfer for anime sketches with enhanced residual u-net and auxiliary classifier gan,'' in \emph{2017 4th IAPR Asian conference on pattern recognition (ACPR)}.\hskip 1em plus 0.5em minus 0.4em\relax IEEE, 2017, pp. 506--511.

\bibitem{wang2023stylediffusion}
Z.~Wang, L.~Zhao, and W.~Xing, ``Stylediffusion: Controllable disentangled style transfer via diffusion models,'' in \emph{Proceedings of the IEEE/CVF International Conference on Computer Vision}, 2023, pp. 7677--7689.

\bibitem{li2021weather}
X.~Li, K.~Kou, and B.~Zhao, ``Weather gan: Multi-domain weather translation using generative adversarial networks,'' \emph{arXiv preprint arXiv:2103.05422}, 2021.

\bibitem{qian2024weatherdg}
C.~Qian, Y.~Guo, Y.~Mo, and W.~Li, ``Weatherdg: Llm-assisted procedural weather generation for domain-generalized semantic segmentation,'' \emph{arXiv preprint arXiv:2410.12075}, 2024.

\bibitem{ma2024deepcache}
X.~Ma, G.~Fang, and X.~Wang, ``Deepcache: Accelerating diffusion models for free,'' in \emph{Proceedings of the IEEE/CVF conference on computer vision and pattern recognition}, 2024, pp. 15\,762--15\,772.

\bibitem{chen2023speed}
Y.-H. Chen, R.~Sarokin, J.~Lee, J.~Tang, C.-L. Chang, A.~Kulik, and M.~Grundmann, ``Speed is all you need: On-device acceleration of large diffusion models via gpu-aware optimizations,'' in \emph{Proceedings of the IEEE/CVF Conference on Computer Vision and Pattern Recognition}, 2023, pp. 4651--4655.

\bibitem{lv2024fastercache}
Z.~Lv, C.~Si, J.~Song, Z.~Yang, Y.~Qiao, Z.~Liu, and K.-Y.~K. Wong, ``Fastercache: Training-free video diffusion model acceleration with high quality,'' \emph{arXiv preprint arXiv:2410.19355}, 2024.

\bibitem{li2023climatenerf}
Y.~Li, Z.-H. Lin, D.~Forsyth, J.-B. Huang, and S.~Wang, ``Climatenerf: Extreme weather synthesis in neural radiance field,'' in \emph{Proceedings of the IEEE/CVF International Conference on Computer Vision}, 2023, pp. 3227--3238.

\bibitem{chen2025styledstreets}
Y.~Chen, Y.~Wang, X.~Zhang, K.~Zhan, P.~Jia, Y.~Zhan, and X.~Lang, ``Styledstreets: Multi-style street simulator with spatial and temporal consistency,'' \emph{arXiv preprint arXiv:2503.21104}, 2025.

\bibitem{yang2024depth}
L.~Yang, B.~Kang, Z.~Huang, X.~Xu, J.~Feng, and H.~Zhao, ``Depth anything: Unleashing the power of large-scale unlabeled data,'' in \emph{Proceedings of the IEEE/CVF Conference on Computer Vision and Pattern Recognition}, 2024, pp. 10\,371--10\,381.

\bibitem{li2024dngaussian}
J.~Li, J.~Zhang, X.~Bai, J.~Zheng, X.~Ning, J.~Zhou, and L.~Gu, ``Dngaussian: Optimizing sparse-view 3d gaussian radiance fields with global-local depth normalization,'' in \emph{Proceedings of the IEEE/CVF Conference on Computer Vision and Pattern Recognition}, 2024, pp. 20\,775--20\,785.

\bibitem{barron2022mip}
J.~T. Barron, B.~Mildenhall, D.~Verbin, P.~P. Srinivasan, and P.~Hedman, ``Mip-nerf 360: Unbounded anti-aliased neural radiance fields,'' in \emph{Proceedings of the IEEE/CVF conference on computer vision and pattern recognition}, 2022, pp. 5470--5479.

\bibitem{Bijelic_2020_CVPR}
M.~Bijelic, T.~Gruber, F.~Mannan, F.~Kraus, W.~Ritter, K.~Dietmayer, and F.~Heide, ``Seeing through fog without seeing fog: Deep multimodal sensor fusion in unseen adverse weather,'' in \emph{The IEEE/CVF Conference on Computer Vision and Pattern Recognition (CVPR)}, June 2020.

\bibitem{jin2022structure}
Y.~Jin, W.~Yan, W.~Yang, and R.~T. Tan, ``Structure representation network and uncertainty feedback learning for dense non-uniform fog removal,'' in \emph{Asian Conference on Computer Vision}.\hskip 1em plus 0.5em minus 0.4em\relax Springer, 2022, pp. 155--172.

\bibitem{li2019benchmarking}
B.~Li, W.~Ren, D.~Fu, D.~Tao, D.~Feng, W.~Zeng, and Z.~Wang, ``Benchmarking single-image dehazing and beyond,'' \emph{IEEE Transactions on Image Processing}, vol.~28, no.~1, pp. 492--505, 2019.

\bibitem{li2019single}
S.~Li, I.~B. Araujo, W.~Ren, Z.~Wang, E.~K. Tokuda, R.~H. Junior, R.~Cesar-Junior, J.~Zhang, X.~Guo, and X.~Cao, ``Single image deraining: A comprehensive benchmark analysis,'' in \emph{Proceedings of the IEEE/CVF Conference on Computer Vision and Pattern Recognition}, 2019, pp. 3838--3847.

\bibitem{schmidt2021climategan}
V.~Schmidt, A.~S. Luccioni, M.~Teng, T.~Zhang, A.~Reynaud, S.~Raghupathi, G.~Cosne, A.~Juraver, V.~Vardanyan, A.~Hernandez-Garcia \emph{et~al.}, ``Climategan: Raising climate change awareness by generating images of floods,'' \emph{arXiv preprint arXiv:2110.02871}, 2021.

\bibitem{rombach2022high}
R.~Rombach, A.~Blattmann, D.~Lorenz, P.~Esser, and B.~Ommer, ``High-resolution image synthesis with latent diffusion models,'' in \emph{Proceedings of the IEEE/CVF conference on computer vision and pattern recognition}, 2022, pp. 10\,684--10\,695.

\bibitem{zhang2018density}
H.~Zhang and V.~M. Patel, ``Density-aware single image de-raining using a multi-stream dense network,'' in \emph{Proceedings of the IEEE conference on computer vision and pattern recognition}, 2018, pp. 695--704.

\bibitem{jayasumana2024rethinking}
S.~Jayasumana, S.~Ramalingam, A.~Veit, D.~Glasner, A.~Chakrabarti, and S.~Kumar, ``Rethinking fid: Towards a better evaluation metric for image generation,'' in \emph{Proceedings of the IEEE/CVF Conference on Computer Vision and Pattern Recognition}, 2024, pp. 9307--9315.

\bibitem{wang2021rich}
Y.~Wang, J.~Ke, H.~Talebi, J.~G. Yim, N.~Birkbeck, B.~Adsumilli, P.~Milanfar, and F.~Yang, ``Rich features for perceptual quality assessment of ugc videos,'' in \emph{Proceedings of the IEEE/CVF conference on computer vision and pattern recognition}, 2021, pp. 13\,435--13\,444.

\bibitem{kejriwal2024challenges}
M.~Kejriwal, E.~Kildebeck, R.~Steininger, and A.~Shrivastava, ``Challenges, evaluation and opportunities for open-world learning,'' \emph{Nature Machine Intelligence}, vol.~6, no.~6, pp. 580--588, 2024.

\end{thebibliography}

\vfill

\end{document}